\documentclass[twocolumn]{article}
\usepackage{natbib}
\usepackage{moreverb,url}
\usepackage{authblk}
\usepackage[colorlinks,bookmarksopen,bookmarksnumbered,citecolor=red,urlcolor=red]{hyperref}
\usepackage{graphicx}
\usepackage{amsmath,amssymb}
\usepackage{caption}
\usepackage{subcaption}
\usepackage{float}
\usepackage{array}
\usepackage{multirow}
\usepackage{booktabs}

\newcolumntype{P}[1]{>{\centering\arraybackslash}p{#1}}
\newcommand{\ra}[1]{\renewcommand{\arraystretch}{#1}}
\newcommand\BibTeX{{\rmfamily B\kern-.05em \textsc{i\kern-.025em b}\kern-.08em
T\kern-.1667em\lower.7ex\hbox{E}\kern-.125emX}}
\newcommand{\keywords}[1]{\small\textbf{\textit{Keywords---}} #1}

\title{Learning Manipulation under Physics Constraints with Visual Perception}

\author[1]{Wenbin Li}
\author[2]{Ale\v{s} Leonardis}
\author[3]{Jeannette Bohg} 
\author[1]{Mario Fritz}
\affil[1]{Max Planck Institute for Informatics, Saarland Informatics Campus, Germany}
\affil[2]{School of Computer Science, University of Birmingham, UK}
\affil[3]{Department of Computer Science, Stanford University, USA}
\date{}

\begin{document}
\maketitle

\begin{abstract}
Understanding physical phenomena is a key competence that enables humans and animals to act and interact under uncertain perception in previously unseen environments containing novel objects and their configurations. In this work, we consider the problem of autonomous block stacking and explore solutions to learning manipulation under physics constraints with visual perception inherent to the task. Inspired by the intuitive physics in humans, we first present an end-to-end learning-based approach to predict stability directly from appearance, contrasting a more traditional model-based approach with explicit 3D representations and physical simulation. We study the model's behavior together with an accompanied human subject test. It is then integrated into a real-world robotic system to guide the placement of a single wood block into the scene without collapsing existing tower structure. To further automate the process of consecutive blocks stacking, we present an alternative approach where the model learns the physics constraint through the interaction with the environment, bypassing the dedicated physics learning as in the former part of this work. In particular, we are interested in the type of tasks that require the agent to reach a given goal state that may be different for every new trial. Thereby we propose a deep reinforcement learning framework that learns policies for stacking tasks which are parametrized by a target structure. 
\end{abstract}


%----------------------------------------------------------------------------
\section{Introduction}
Understanding and predicting physical phenomena in daily life is an important component of human intelligence. This ability enables us to effortlessly manipulate objects in previously unseen conditions. It is an open question how this kind of knowledge can be represented and what kind of models could explain human manipulation behavior~\citep{yildirim2017physical}. In this work, we explore potential frameworks for the robot to learn manipulation with respect to the corresponding physics constraints underneath the task. 

Behind the human's physics reasoning in everyday life, the \textit{intuitive physics} \citep{Smith1994, mccloskey1983intuitive} plays an important role in the process, representing \textit{raw} knowledge for human to understand the physical environment and interactions. Albeit sometimes erroneous, it works well enough for most situations in daily life. It has been an ongoing research in cognitive science and psychology, among others, to understand computational models~\citep{battaglia2012computational} and explain such a mechanism. 

It has not yet been shown how to equip machines with a similar set of physics commonsense---thereby bypassing a model-based representation and a physical simulation. In fact, it has been argued that such an approach is unlikely due to e.g., the complexity of the problem~\citep{battaglia2013simulation}. Only recently, several works have revived this idea and reattempted a fully data driven approach to capturing the essence of physical events via machine learning methods \citep{mottaghi2015newtonian,wu2015galileo,fragkiadaki2015learning,apratim16ariv}. 

In the first part of this work, we draw inspiration from the studies in developmental psychology~\citep{baillargeon1994infants,baillargeon2008innate} where the infants acquire the knowledge of physical events gradually through the observation of various event instances. In this context, we revisit the classical setup of Tenenbaum and colleagues~\citep{battaglia2013simulation} and explore to which extent machines can predict physical stability events directly from appearance cues. We approach this problem by synthetically generating a large set of wood block towers under a range of conditions, including varying number of blocks, varying block sizes, planar vs.\ multi-layered configurations. We run those configurations through a simulator ({\em only at training time!}) in order to generate labels whether a tower would fall or not. We show for the first time that the aforementioned stability test can be learned and predicted in a purely data driven 
way---bypassing traditional model-based simulation approaches. Further, we accompany our experimental study with human judgments on the same stimuli. Then we utilize this framework to guide the robot to stably stack a single block onto the existing block-structure based on the stability prediction. To circumvent the domain shift between the synthesized images and the real world scene images, we extract the foreground masks for both synthesized and captured images. Given a real world block structure, the robot uses the model trained on the synthesized data to predict the stability outcome across possible candidate placements, and performs stacking on the feasible locations afterwards. We evaluate both the prediction and manipulation performance on the very task.

\begin{figure}
\centering
\includegraphics[width=0.68\linewidth]{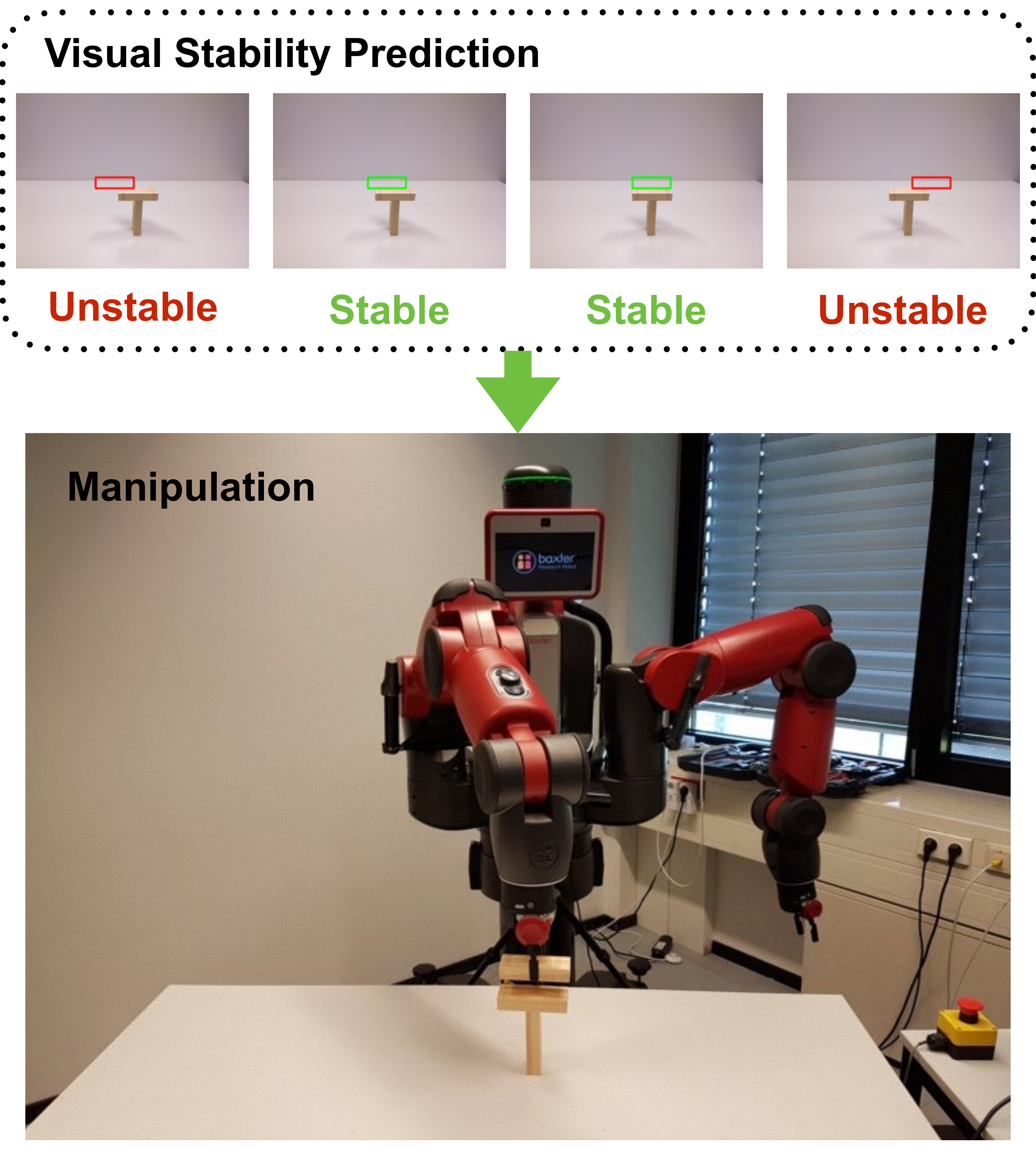}
\caption{Given a wood block structure, our visual stability classifier predicts the stability for future placements, and the robot then stacks a block among the predicted stable placements.}
\label{fig:teaser}
\end{figure}

In the second part of this work, we further tackle a more challenging stacking task (target stacking). The goal for the task is to reproduce a shape shown in an image by consecutively stacking multiple blocks while retaining physical stability and avoiding pre-mature collisions with the existing structure. We explore an alternative modeling to learn the block stacking through trial-and-error, bypassing the need to explicitly model the corresponding physics knowledge as in the former part of this work. 
For this purpose, we build a synthetic environment with physics simulation, where the agent can move and stack blocks and observe the different outcomes of its actions. We apply deep reinforcement learning to directly acquire the block stacking skill in an end-to-end fashion. By introducing the goal-parameterized policies, we learn a single model to guide the agent to build different shapes on request. We first validated this model on a toy example where the agent has to navigate in a grid-world where both the location of the start and end point are randomized for each episode and then the target stacking. On both experiments, we observe generalization across different goals.

We tackle the block stacking task, by focusing the manipulation learning with respect to the corresponding physics constraints underneath the task. The first part of the work emphasizes the visual stability prediction formalism and its application to single block stacking task that was initially presented in~\cite{li2017visual}. The second part of the work features an extension to  goal-parametrized deep reinforcement learning framework and its application to target stacking task. In addition, we now include a discussion of the simulation environment implemented for both tasks. While simulation engines and game engines, such as Panda3D \citep{goslin2004panda3d}, provide a generic platform to devise environment to different needs, it is still very time-consuming and remains non-trivial to customize one for a specific tasks such as the target stacking in this work. Hence, we briefly recap our design for the environments to provide readers more insights for further practice. 
\section{Related Work}
\subsection{Physics Understanding}
Humans possess the amazing ability to perceive and understand ubiquitous physical phenomena occurring in their daily life. This gives rise to the concept of intuitive physics, aiming to describe the knowledge which enables humans to understand physical environment and interact accordingly. In particular, the ``intuitive'' part emphasizes knowledge which is considered commonsense to ordinary people not reliant on specialized training in physics. Intuitive physics is ubiquitous in guiding humans' actions in daily life, such as where to put a cup stably and how to catch a ball. Along the years, research on intuitive physics has been conducted from many different perspectives across psychology, cognitive science and artificial intelligence.

In developmental psychology, researchers seek to understand how this ability develops. \cite{baillargeon2002acquisition} suggest that infants acquire the knowledge of physical events at a very young age by observing those events, including support events and others. 
Interestingly, in a recent work \cite{denil2016learning}, the authors introduce a basic set of tasks that require the learning agent to estimate physical properties (mass and cohesion combinations) of objects in an interactive simulated environment and find that it can learn to perform the experiments strategically to discover such hidden properties in analogy to human's development of physics knowledge.
Another interesting question that has been explored in psychology is how knowledge about physical events affects and guides human's actual interaction with objects \cite{yildirim2017physical}. 
Yet it is not clear how a machine model trained for physics understanding can directly be applied to real-world interactions with objects and accomplish manipulation tasks.

In cognitive science, \cite{battaglia2013simulation} proposes an intuitive physics simulation engine as an internal mechanism for such type of ability and found close correlation between its behavior patterns and human subjects' on several psychological tasks.

More recently, there has been renewed interest in physics understanding in computer vision and machine learning communities. For instance, understanding physical events plays an important role in scene understanding in computer vision. By including additional clues from physical constraints into the inference mechanism, mostly from the support event, it has further improved results in segmentation of surfaces \cite{gupta2010blocks}, scenes \cite{silberman2012indoor} from image data, and object segmentation in 3D point cloud data \cite{zheng2013beyond}. 
Another research direction is to equip artificial agents with such an ability by letting them learn physical concepts from visual data. 
\cite{mottaghi2015newtonian} aim at understanding dynamic events governed by laws of Newtonian physics and use proto-typical motion scenarios as exemplars. \cite{fragkiadaki2015learning} analyze billiard table scenarios and learn dynamics from observation with explicit object notion. 
An alternative approach based on boundary extrapolation \cite{apratim16ariv} addresses similar settings without imposing any object notion. 
\cite{wu2015galileo} aims to understand physical properties of objects based on explicit physical simulation. 
\cite{mottaghi2017see} proposes to reason about containers and the behavior of the liquids inside them from a single RGB image.

A most related work is done by \cite{fergus16blocsarxiv}, where the authors propose using a visual model to predict stability and falling trajectories for simple 4 block scenes. In the first part of this work, we investigate if and how the prediction performance of such image-based models changes when trained on block stacking scenes with larger variety and further examine how the human's prediction adapts to the variation in the generated scenes and compare to the learned visual model. Each work requires significant amounts of simulated, physically-realistic data to train the large-capacity, deep models.

\subsection{Learning from Synthetic Data and Simulation}
Learning from synthetic data has a long tradition in computer vision and has recently gained increasing interest \citep{li12eccv,kostas14cvpr,peng2015learning,rematas16cvpr} due to data hungry deep-learning approaches. 
In the first part of this work, we use a game engine to render scene images and a built-in physics simulator to simulate the scenes' stability behavior. The data generation procedure is based on the platform used in \cite{battaglia2013simulation}, however as discussed before, their work hypothesized a simulation engine as an internal mechanism for human to understand the physics in the external world while we are interested in finding an image-based model to directly predict the physical behavior from visual channel. 

In reinforcement learning domain, it is a common practice to utilize simulation environment to train the reinforcement agent, such as the Atari Games~\citep{bellemare13arcade} and ViZDoom~\citep{kempka2016vizdoom}. However, there is no off-the-shelf environment supporting trial-and-error interaction for block stacking. Hence, in the second part of this work, we build on existing game engine and create our interactive environment with physics simulation to allow the learning agent to learn stacking skills through interactions.

\subsection{Blocks-based Manipulation Tasks}
To shed more light on the capabilities of our model, in the first part of this work, we explore how the visual stability prediction model can be used in a robotic manipulation task, i.e., stably stacking a wood block given a block structure. In the past, we have seen researchers perform tasks with wood blocks, like playing Jenga from different perspectives. \cite{kroger2006demonstration} demonstrated multi-sensor integration by using a marker-based system with multiple cameras and sensors: a random block is first chosen in the tower, then the robot arm will try to pull the very block, if the force sensor detects large counter force or the CCD cameras detect large motion of tower, then it will stop pulling and try other block. 

\cite{wang2009robot} improved on \cite{kroger2006demonstration} by further incorporating a physics engine to initialize the candidates for pulling test. In comparison, we do not fit 3D models or run physics simulation at test time for the given scene but instead use the scene image as input to directly predict the physics of the structure. 

A different line of research is \cite{kimura2010force} where physical force is explicitly formulated with respect to the tower structure for planning. In our work, we do not do explicit formation of contact force as in \cite{kimura2010force}, nor do we perform trials on-site for evaluating the robot's operation. We only use physics engine to acquire synthesized data for training the visual-physics model. At test time, the planning system for our robot mainly exploits the knowledge encoded in the visual-physics model to evaluate the feasibility of individual candidates and performs operations accordingly. 

\subsection{Reinforcement Learning}
In the second part of this work, reinforcement learning is used to learn an end-to-end model directly from the experience collected during interaction with a physically-realistic environment.
The majority of work in reinforcement learning focuses on solving task with a single goal. However, there are also tasks where the goal may change for every trial. It is not obvious how to directly apply the model learned towards a specific goal to a different one. An early idea has been proposed by \cite{kaelbling1993learning} for a maze navigation problem in which the goal changes. The author introduces an analogous formulation to the Q-learning by using shortest path in replacement of the value functions. 
Yet there are two major limitations for the framework: 1) it is only formulated in tabular form which is not practical for application with complex states 2) the introduced shortest path is very specific to the maze navigation setting and hence cannot be easily adapt to handle task like target stacking. 
In contrast, we propose a goal-parameterized model to integrate goal information into a general learning-based framework that facilitates generalization across different goals. The model has been shown to work on both a navigation task and target stacking.

A few other works also explored the idea of integrating goal information into learning. \cite{schaul2015universal} propose the universal value function approximators (UVFAs) to integrate goal information into learning. \cite{oh2017zero} proposes zero-shot task generation with explicit modeling of skills and subtasks where our approach is end-to-end and thereby bypasses this kind of modeling.
\cite{dosovitskiy2016learning} use the terms ``goal'' and ``target''. However, their notion is different from ours. Their framework is using auxiliary information (intermediate measurements as goal module) to improve learning to maximize gaming score in visually-richer task like vizDoom. There is no specific goal to enforce the agent to reach certain state which is exactly what our approach facilitates.
The metacontroller \citep{hamrick2017metacontrol}, the imagination strategy \citep{pascanu2017learning} and the I2A model \citep{weber2017imagination} are all appealing generic frameworks to boost overall performance over existing RL methods. While these approaches -- similar to some of the Atari games -- have a goal encoded in the observation/state, this goal remains implicit. This is applicable in many robotic scenarios, where goals are specified externally. 
Therefore, we explore an approach for target stacking that has an explicit notion and model of goals. Our work is the first to bring this concept to bear towards manipulation and planing under physical constraints -- breaking with more conventional simulation and planning approaches (e.g. \cite{yildirim2017physical}).
\section{Simulation Environment}
Modern machine learning techniques build on data. With the rise of deep learning models, the need of data becomes even more prominent. One of the key elements for the recent success of deep learning model in domains like generic image and speech recognition is the abundance of data in related fields. However, when it comes to a specific domain, it often runs short of data and in reality, collecting data is still an expensive process. One remedy for this issue is to exploit the domain knowledge to synthesize data and utilize the generated data for learning. This is also the case for our study as there is no obvious source of clean and sufficient data to learn physics from. Hence, we make use of the Panda3D~\footnote{\url{https://www.panda3d.org/}} to create both simulation environments in our work. 

Panda3D is an open source game engine for Python and C++ programs. It was originally developed by the Disney's VR studio to be ``flexible enough to support everything from realtime graphics applications to the development of high-end virtual reality theme park attractions or video games'' \citep{goslin2004panda3d} and it has evolved significantly along the years. 

As a game engine, the Panda3D provides additional capabilities besides 3D rendering including the physics system with integration of different physics engines. Asides from its own built-in basic physics engine, it also supports more advanced ones including the Open Dynamics Engine (ODE) \footnote{\url{http://www.ode.org/}} \citep{smith2005open} and the Bullet \footnote{\url{http://bulletphysics.org/wordpress/}} \citep{coumans2010bullet} for physics simulation back-end. We used the Bullet in both of our simulation environments.

\begin{figure}
\centering
\includegraphics[width=0.97\linewidth,keepaspectratio]{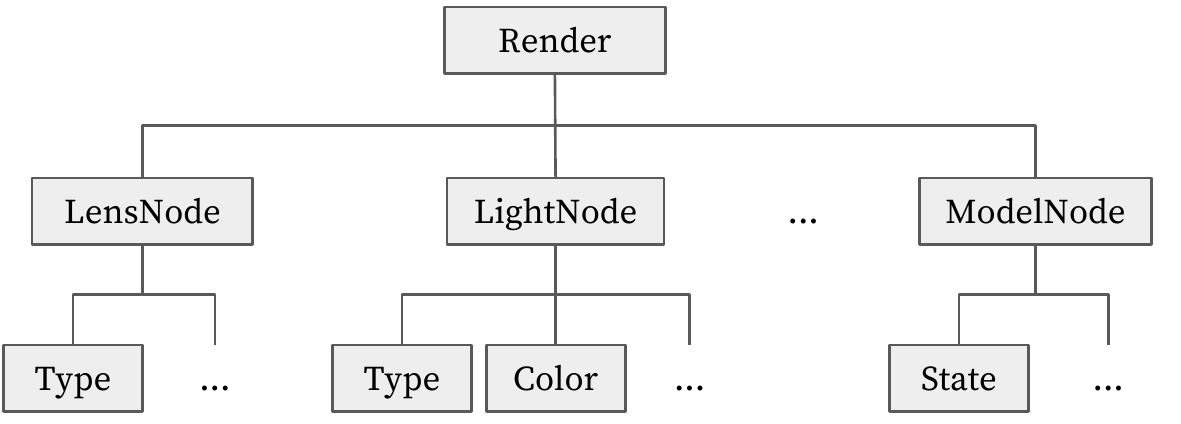}
\caption{An example of scene graph in Panda3D.}
\label{fig:sim:panda3d_sceneGraph}
\end{figure}

The workflow of Panda3D builds on the concept of scene graph. The \textbf{Scene graph} is a general data structure to represent and organize a graphical scene. In Panda3D, the scene graph is maintained as a tree of objects to be rendered. The tree consists of objects of class \texttt{PandaNode}. As shown in Figure~\ref{fig:sim:panda3d_sceneGraph}, the root node is called the \texttt{render} and the rest define different perspectives of the scene with various attributes. For example, the \texttt{LensNode} controls the camera such as the perspective, the \texttt{LightNode} manages the lighting in the scene such as the color and type of the lighting and the \texttt{ModelNode} encodes the 3D object in the frame.

\begin{figure}
\centering
\includegraphics[width=0.78\linewidth,keepaspectratio]{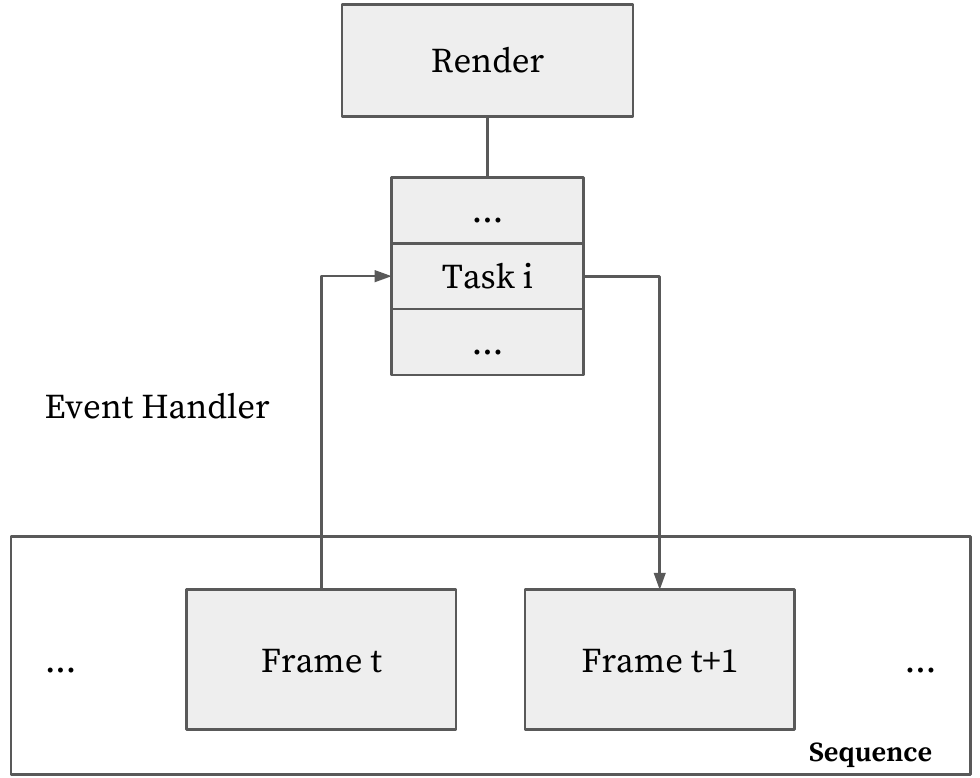}
\caption{The render process in Panda3D.}
\label{fig:sim:panda3d_render}
\end{figure}

Another important concept is the \textbf{task}. Tasks are functions called by Panda3D at every frame or for every specified amount of time. Together with \textbf{event handler} which is called upon special conditions (\textbf{events}) occur, update can be made to the scene in Panda3D between rendering steps as shown in Figure~\ref{fig:sim:panda3d_render}. For instance, the task can be the simulation subroutine that updates the states of objects in the scene caused by physics. 

\section{Part I: From Visual Stability Prediction to Single Block Stacking}

\subsection{Stability Prediction from Still Images}
\paragraph{Inspiration from Human Studies}
Research in \cite{hamrick2011internal,battaglia2013simulation} suggests the combinations of the most salient features in the scenes are insufficient to capture people's judgments, however, contemporary study reveals human's perception of visual information, in particular some geometric feature, like critical angle \cite{cholewiak2013visual,cholewiak2015perception} plays an important role in the process. Regardless of the actual inner mechanism for humans to parse the visual input, it is clear there is a mapping $f$ involving visual input $I$ to the stability prediction $P$. 
\begin{gather*}
\begin{aligned}
f: I,\ast \rightarrow P
\end{aligned}
\end{gather*}
Here, $\ast$ denotes other possible information, i.e., the mapping can be inclusive, as in \cite{hamrick2011internal} using it along with other aspects, like physical constraint to make judgment or the mapping is exclusive, as in \cite{cholewiak2013visual} using visual cues alone to decide.

\begin{figure}
\centering
\includegraphics[width=0.95\linewidth]{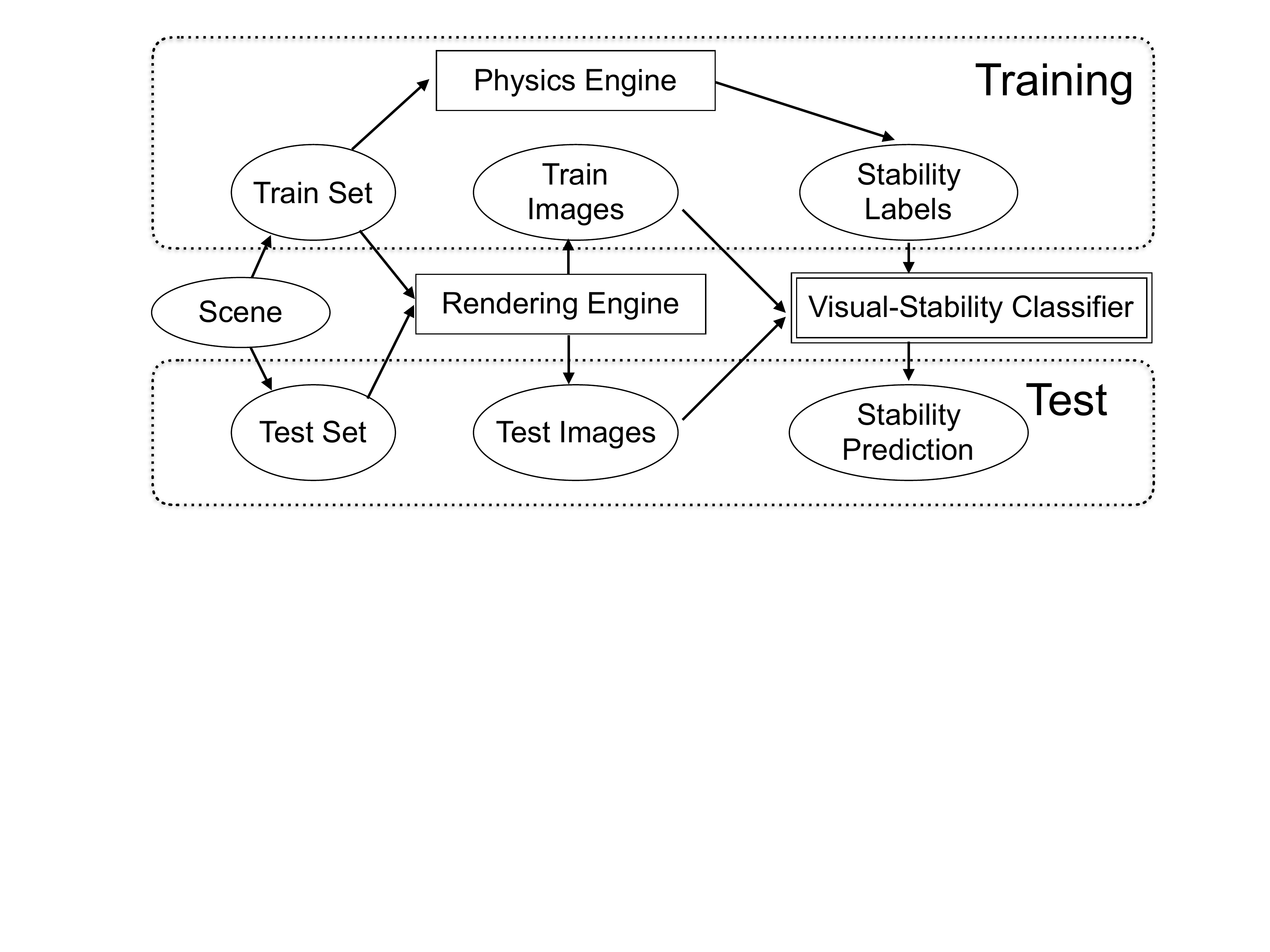}
\caption{An overview of our approach for learning visual stability. Note that physics engine is only used during training time to get the ground truth to train the deep neural network while at test time, only rendered scene images are given to the learned model to predict the physical stability of the scenes.}
\label{fig:overview_recog}
\end{figure}

\paragraph{Image Classifier for Stability Prediction}
In our work, we are interested in the mapping $f$ exclusive to visual input and directly predicts the physical stability (visual stability test). To this end, we use deep convolutional neural networks as it has shown great success on image classification tasks~\citep{krizhevsky2012imagenet}. Such networks have been shown to be able to adapt to a wide range of classification and prediction task \citep{razavian2014cnn} through re-training or adaptation by fine-tuning.
Therefore, these approaches seem to be adequate methods to study visual prediction on this challenging task with the motivation that by changing conventional image classes labels to stability labels the network can learn ``physical stability salient'' features. By setting up a data generation process that allows us to control various degrees of freedom induced by the problem as well as generation of large quantities of data in a repeatable setting, we can then validate our approach. An overview of our approach is shown in Figure~\ref{fig:overview_recog}.

\subsection{Data Generator for Visual Stability Prediction}
\begin{figure}
\centering
\includegraphics[width=1.\linewidth,keepaspectratio]{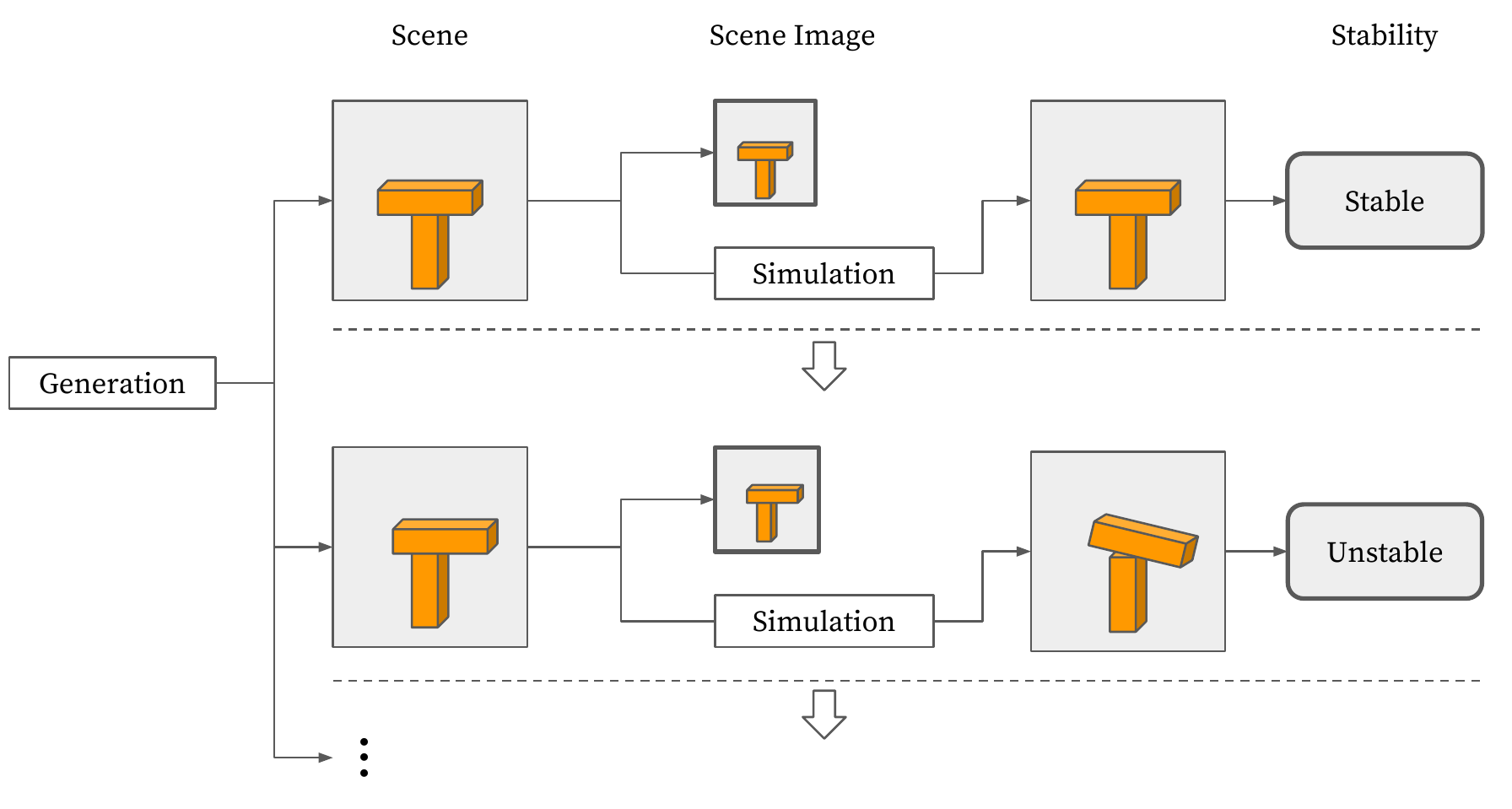}
\caption{Overview of the data generator for visual stability prediction. The scene images and their corresponding stability labels are collected.}
\label{fig:sim:stability_data_pippeline}
\end{figure}

Based on the scene simulation framework used in \citep{hamrick2011internal,battaglia2013simulation}, we build a data generator to synthesize data for visual stability tests. Figure~\ref{fig:sim:stability_data_pippeline} gives an overview of the data generator. The first key component is the \textbf{tower generation}. It automatically generate a large number of different blocks structures (tower) under the \textit{scene parameters}, including the number of blocks, stacking depth and block size (We will elaborate on this later). These towers are recorded as scene files which only track all the locations of blocks in the scene. 

The next component is the \textbf{stability simulation}. It loads the stored scene files and simulates their stability with physics engine. The images for the towers before running the physics engine are captured as the scene images, the stability labels from the simulation are automatically determined and recorded for the corresponding towers. Both the scene images and the obtained stability labels are later put into the deep convolutional neural network to learn the visual stability classifier.

\subsubsection{Tower Generation}
\begin{figure}
\centering
\includegraphics[width=0.65\linewidth,keepaspectratio]{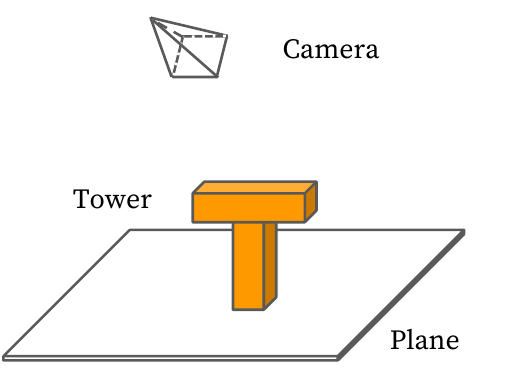}
\caption{Example set-up of the scene for the data generator.}
\label{fig:sim:stability_data_setup}
\end{figure}

An example of the scene setup is shown Figure~\ref{fig:sim:stability_data_setup}. The tower is placed on a plane, and a camera is positioned at the front-facing location of which the elevation is adjusted for the towers of different heights. For different scenes, the towers are generated differently, and the scene images are captured through the camera. To make the scene images more realistic, wood texture is added to all the blocks.

\begin{figure}
\centering
	\begin{subfigure}[b]{0.23\linewidth}
    \centering
    \includegraphics[width=0.98\linewidth]{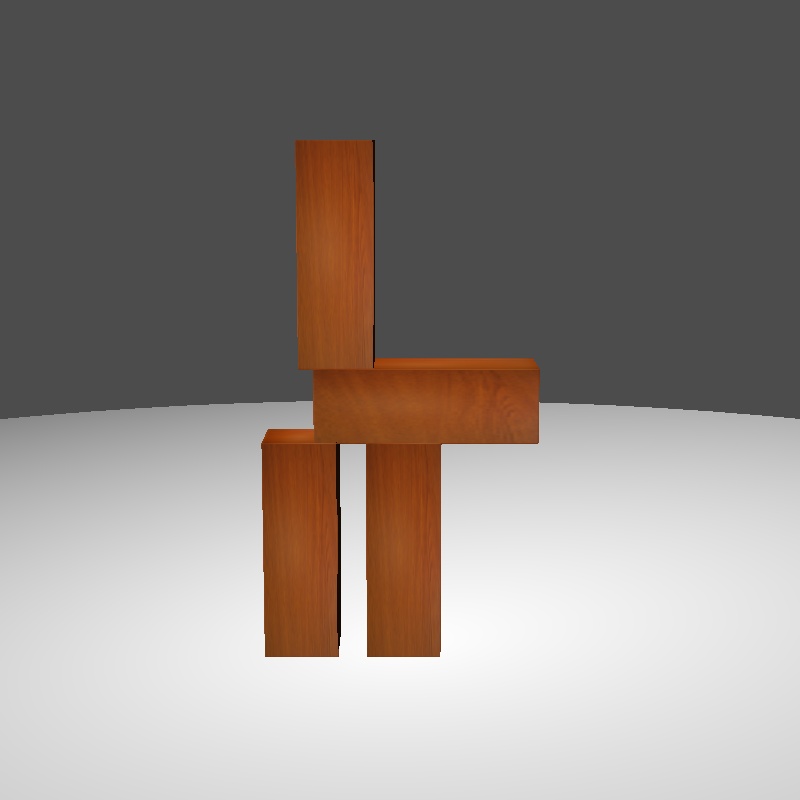}
    \caption{4 Blocks}
    \end{subfigure}
    \begin{subfigure}[b]{0.23\linewidth}
    \centering
    \includegraphics[width=0.98\linewidth]{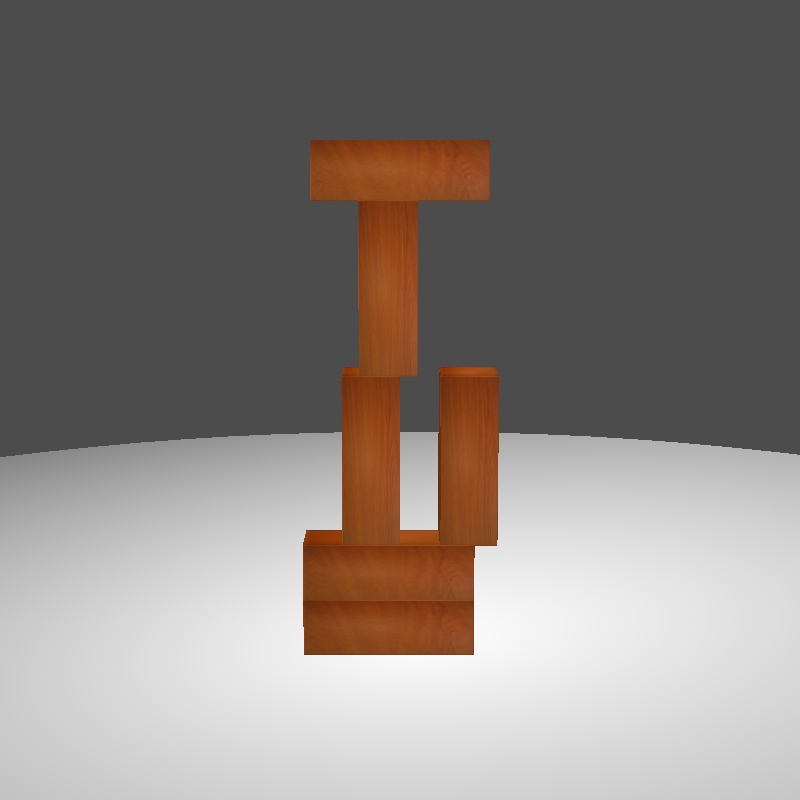}
    \caption{6 Blocks}
    \end{subfigure}
    \begin{subfigure}[b]{0.23\linewidth}
    \centering
    \includegraphics[width=0.98\linewidth]{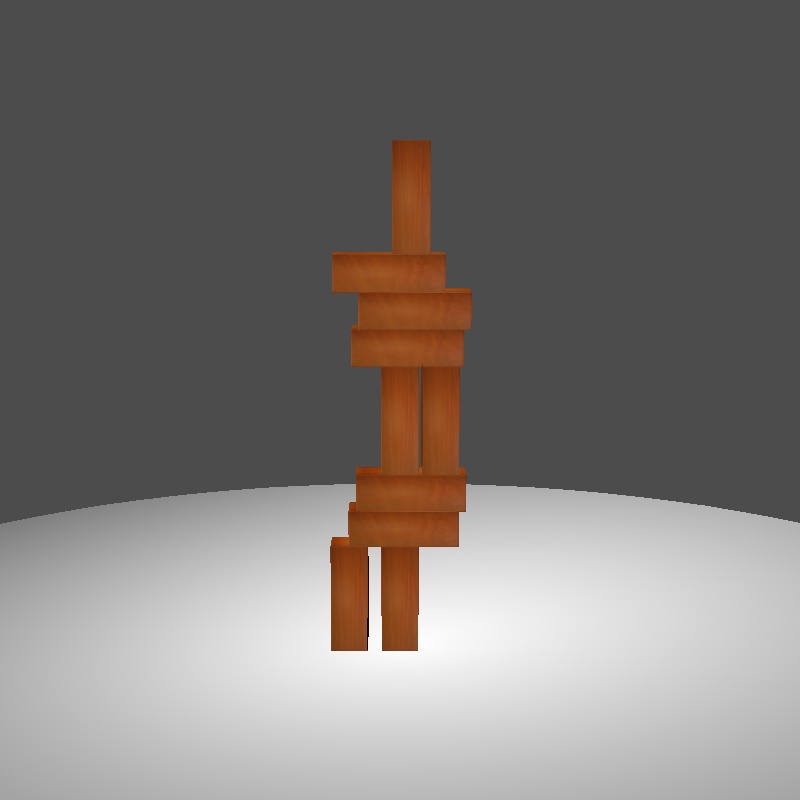}
    \caption{10 Blocks}
    \end{subfigure}
	\begin{subfigure}[b]{0.23\linewidth}
    \centering
    \includegraphics[width=0.98\linewidth]{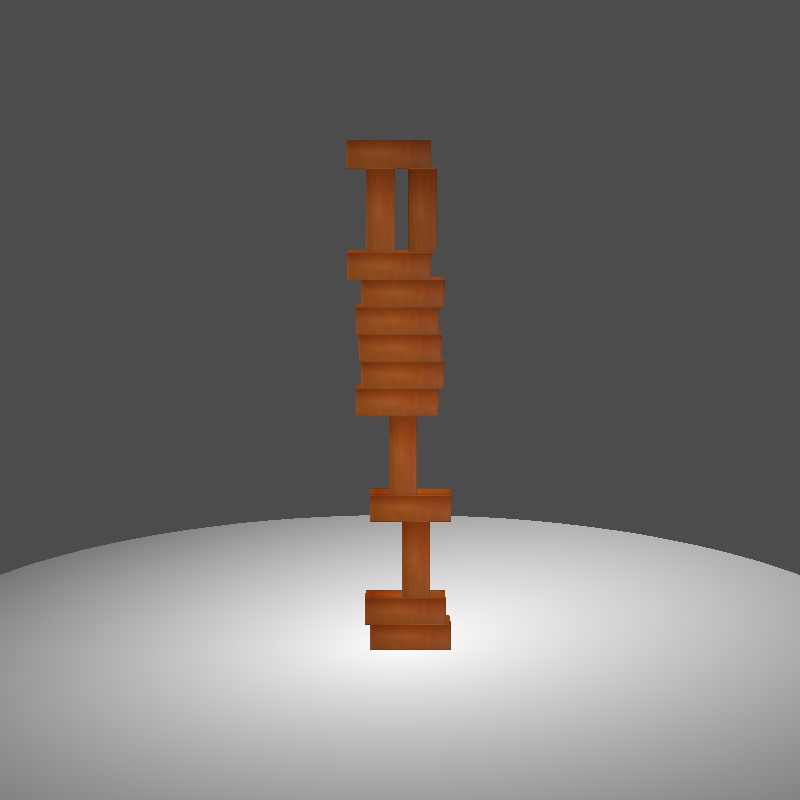}
    \caption{14 Blocks}
    \end{subfigure}
\caption{Example of generated scene images with different total number of blocks.}
\label{fig:sim:stability_example}
\end{figure}

The basic tower generation system is based on the framework by \cite{battaglia2013simulation}. Given a specified number of total blocks in the tower, the blocks are sequentially added into the scene under the geometrical constraints, such as no collision between blocks. Some examples of obtained scene images are shown in Figure~\ref{fig:sim:stability_example}. 

\subsubsection{Stability Simulation}
During the stability simulation, we set the simulation time universally to 2 seconds at $1000{\text Hz}$ for all the scenes. Surface friction and gravity are enabled in the simulation. The system records the configuration of a scene of $N$ blocks at time $t$ as $(p_1,p_2,...,p_N)_t$, where $p_i$ is the location for block $i$. The stability is then automatically decided as a Boolean variable: 
\begin{gather*}
\begin{aligned}
S = \bigvee\limits_{i=1}^N (\Delta((p_i)_{t=T} - (p_i)_{t=0}) > \tau)
\end{aligned}
\end{gather*}
where $T$ is the end time of simulation, $\delta$ measures the displacement for the blocks between the starting point and end time, $\tau$ is the displacement threshold, $\bigvee$ denotes the logical ${\text{Or}}$ operator, that is to say it counts as unstable $S={\text{True}}$ if any block in the scene moved in simulation, otherwise as stable $S={\text {False}}$. This is necessary as the blocks in stable towers can generate small displacement during the simulation process, simply using zero displacement to determine the stability can lead to lots of erroneous cases where the stable towers are labeled as unstable. In practice, we picked the threshold based on the evaluation of a small set of towers.

\subsection{Synthetic Data}
The number of blocks, blocks' size and stacking depth are varied in the generated scenes, to which we will refer as {\it scene parameters}. 
\paragraph{Numbers of Blocks}
We expect that varying the size of the towers will influence the difficulty and challenge the competence of ``eye-balling'' the stability of a tower in humans and machine. While evidently the appearance becomes more complex with the increasing number of blocks, the number of contact surfaces and interactions equally make the problem richer. Therefore, we include scenes with four different number of blocks, i.e., 4 blocks, 6 blocks, 10 blocks and 14 blocks as $\{{\text 4B}, {\text 6B}, {\text 10B}, {\text 14B}\}$.

\paragraph{Stacking Depth}
As we focus our investigations on judging stability from a monocular input, we vary the depth of the tower from a one layer setting which we call ${\text 2D}$ to a multi-layer setting which we call ${\text 3D}$.
The first one only allows a single block along the image plane at all height levels while the other does not enforce such constraint and can expand in the image plane. Visually, the former results in a single-layer stacking similar to Tetris while the latter ends in a multiple-layer structure as shown in Table~\ref{tab:scene_param}. The latter most likely requires the observer to pick up on more subtle visual cues, as many of its layers are heavily occluded. 

\paragraph{Block Size}
We include two groups of block size settings. In the first one, the towers are constructed of blocks that have all the same size of $1 \times 1 \times 3$ as in the \cite{battaglia2013simulation}. The second one introduces varying block sizes where two of the three dimensions are randomly scaled with respect to a truncated Normal distribution $N(1,\sigma^2)$ around $[1 - \delta,1 + \delta]$,  $\sigma$ and $\delta$ are small values. These two settings are referred to as $\{{\text Uni}, {\text NonUni}\}$. The setting with non-uniform blocks introduces small visual cues where stability hinges on small gaps between differently sized blocks that are challenging even for human observers.

\paragraph{Scenes}
Combining these three scene parameters, we define $16$ different scene groups. For example, group 10B-2D-Uni is for scenes stacked with 10 Blocks of same size, stacked within a single layer. For each group, $1000$ candidate scenes are generated where each scene is constructed with non-overlapping geometrical constraint in a bottom-up manner. There are $16 {\text K}$ scenes in total. For prediction experiments, half of the images in each group are for training and the other half for test, the split is fixed across the experiments.

\paragraph{Rendering}
While we keep the rendering basic, we like to point out that we deliberately decided against colored bricks as in \cite{battaglia2013simulation} in order to challenge perception and make identifying brick outlines and configurations more challenging. The lighting is fixed across scenes and the camera is automatically adjusted so that the whole tower is centered in the captured image. Images are rendered at resolution of $800 \times 800$ in color.

\begin{table*}
\begin{tabular}{cccc|cc|cc}
\hline\hline
\multicolumn{4}{c}{Block Numbers} & \multicolumn{2}{c}{Stacking Depth} &\multicolumn{2}{c}{Block Size}\\
\hline
& & & & & & & \\
\begin{subfigure}{0.101\textwidth}\centering\includegraphics[width=0.96\columnwidth]{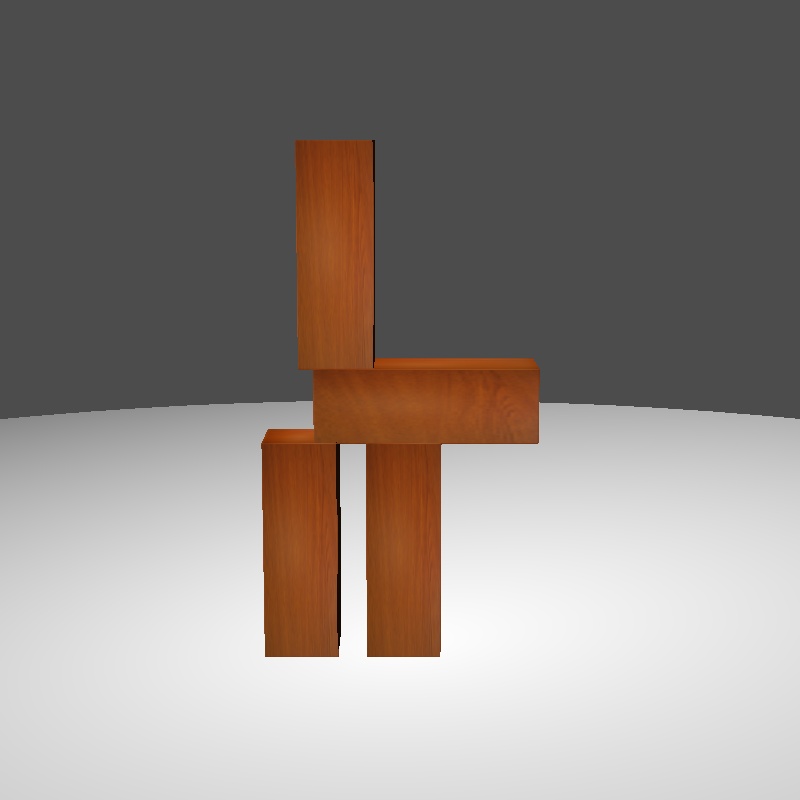}\caption{\scriptsize4 Blocks}\end{subfigure}&
\begin{subfigure}{0.101\textwidth}\centering\includegraphics[width=0.96\columnwidth]{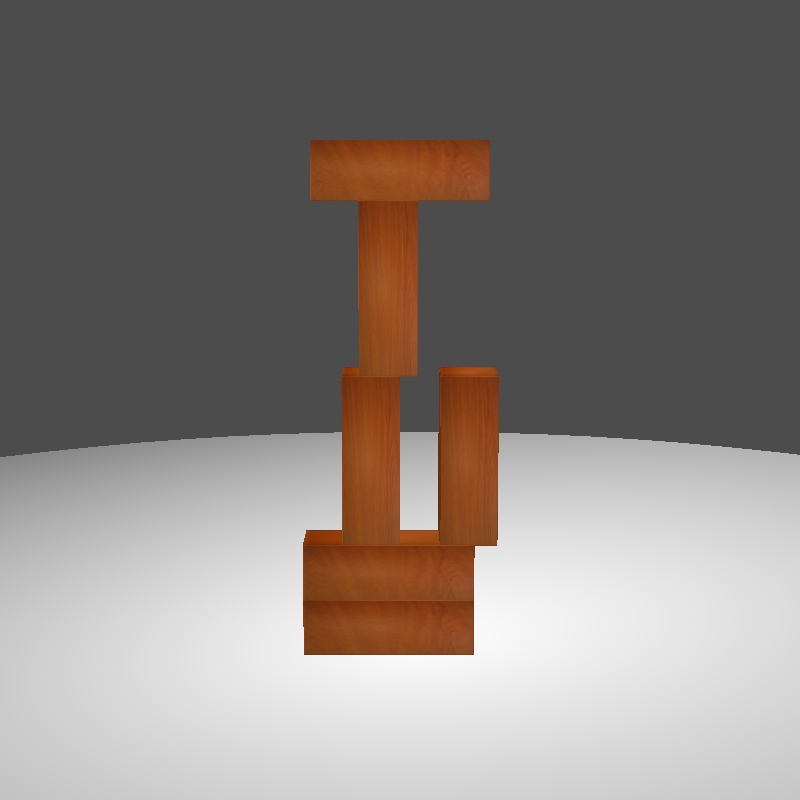}\caption{\scriptsize6 Blocks}\end{subfigure}&
\begin{subfigure}{0.101\textwidth}\centering\includegraphics[width=0.96\columnwidth]{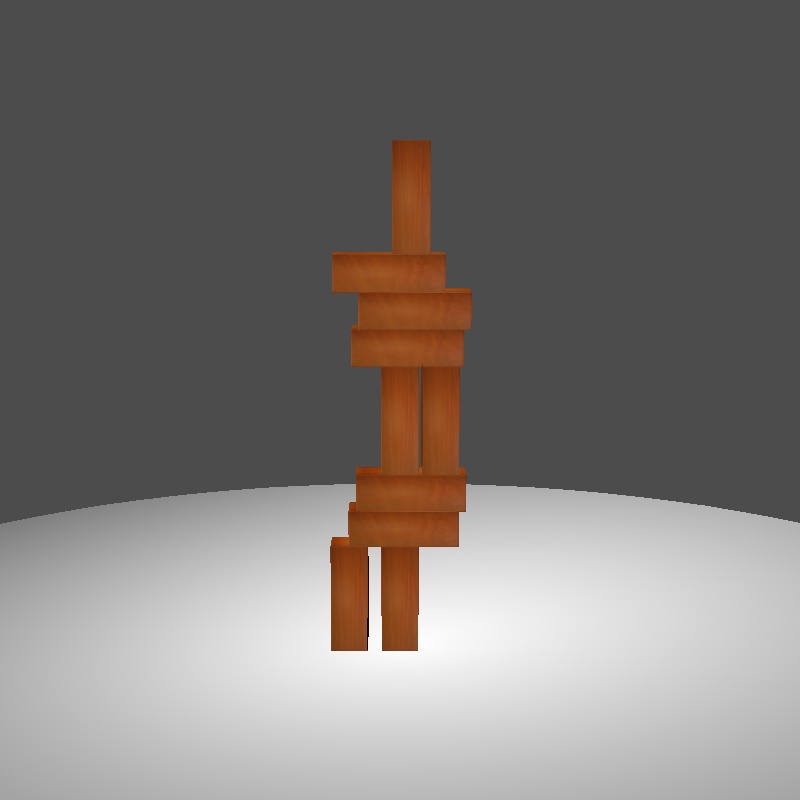}\caption{\scriptsize10 Blocks}\end{subfigure}&
\begin{subfigure}{0.101\textwidth}\centering\includegraphics[width=0.96\columnwidth]{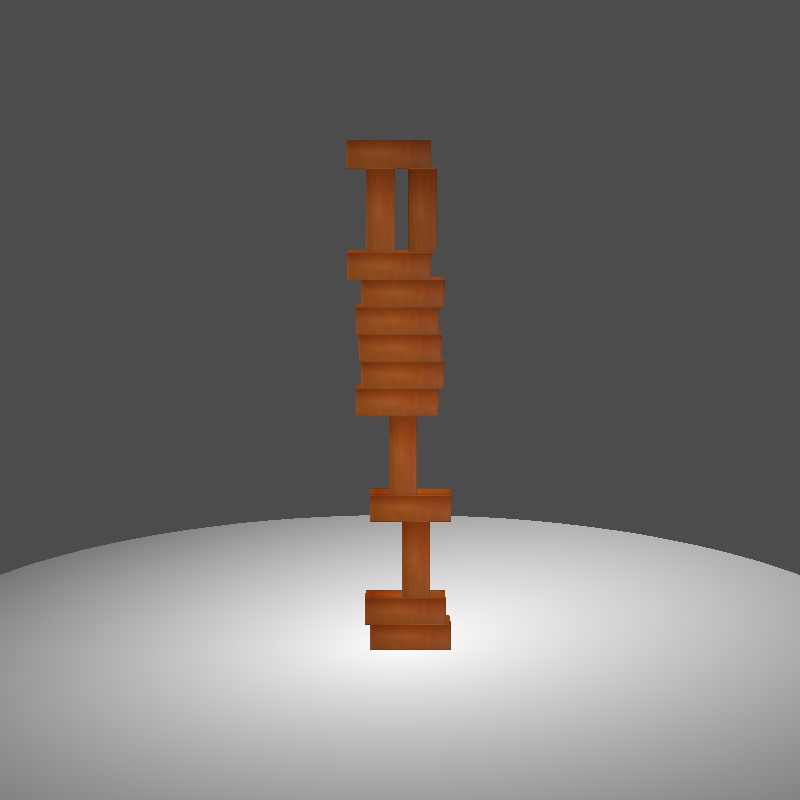}\caption{\scriptsize14 Blocks}\end{subfigure}&
\begin{subfigure}{0.101\textwidth}\centering\includegraphics[width=0.96\columnwidth]{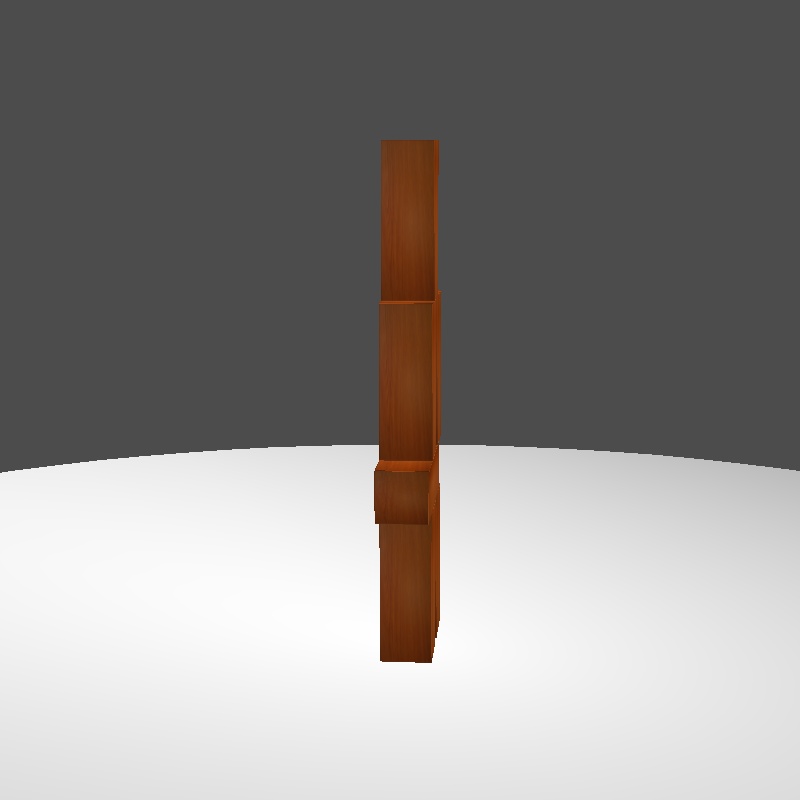}\caption{\scriptsize2D-stack}\end{subfigure}&
\begin{subfigure}{0.101\textwidth}\centering\includegraphics[width=0.96\columnwidth]{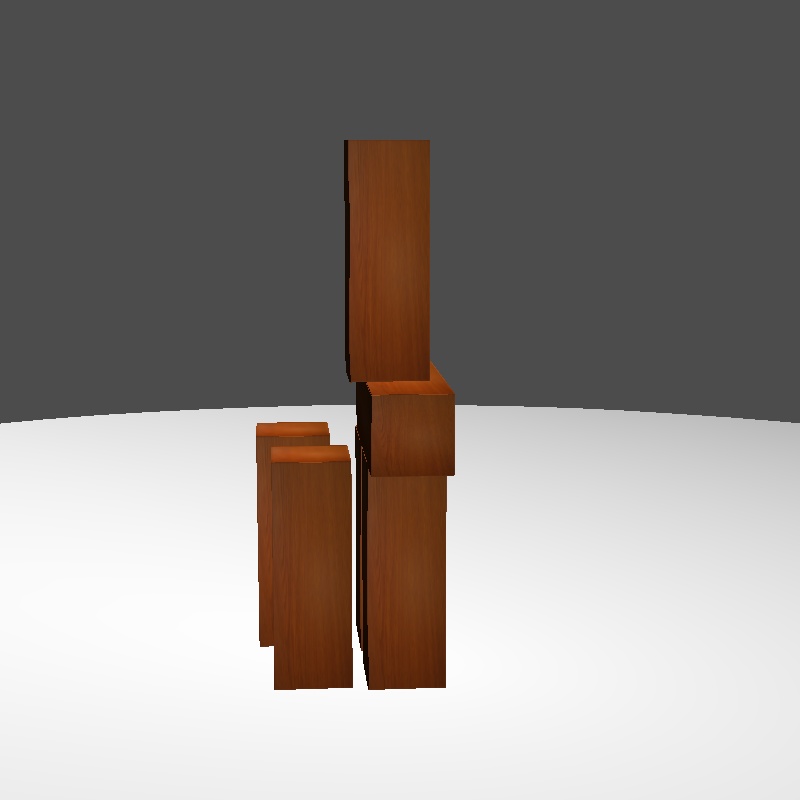}\caption{\scriptsize3D-stack}\end{subfigure}&
\begin{subfigure}{0.101\textwidth}\centering\includegraphics[width=0.96\columnwidth]{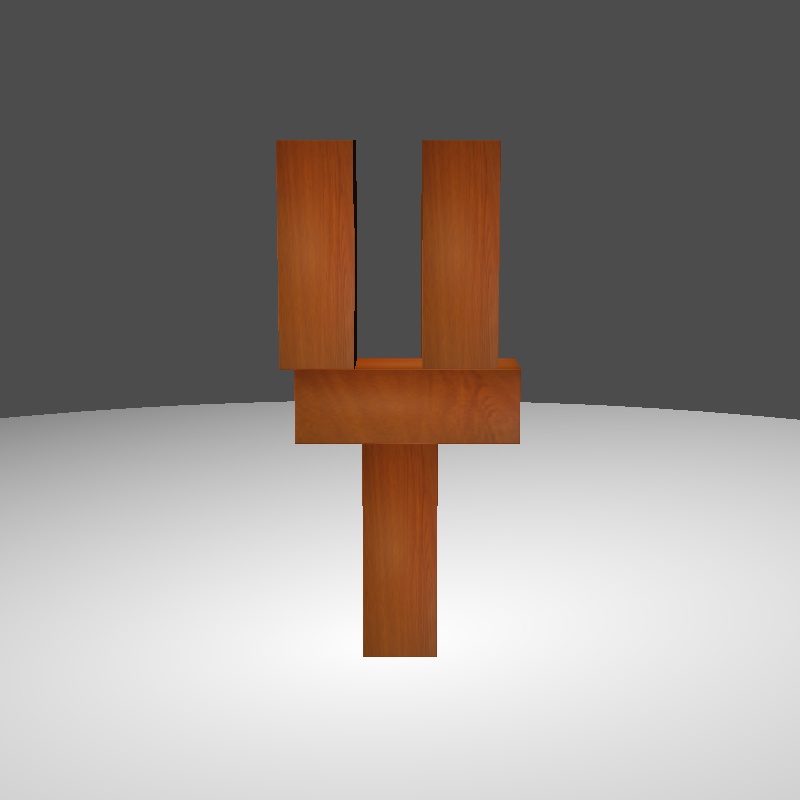}\caption{\scriptsize size-fix}\end{subfigure} &
\begin{subfigure}{0.101\textwidth}\centering\includegraphics[width=0.96\columnwidth]{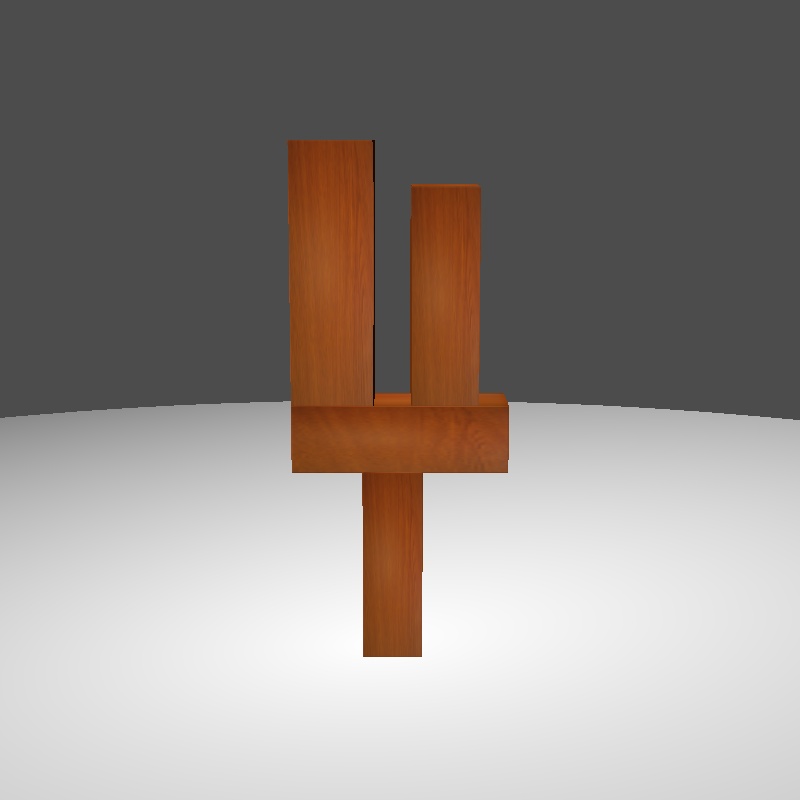}\caption{\scriptsize size-vary}\end{subfigure}\\
\hline
\end{tabular}
\caption{Overview of the scene parameters in our rendered scenes. There are 3 groups of scene parameters across number of blocks, stacking depth and block size.}
\label{tab:scene_param}
\end{table*}

\subsection{Prediction Performance}
In this part of the experiments, the images are captured before the physics engine is enabled, and the stability labels are recorded from the simulation engine as described before. At the training time, the model has access to the images and the stability labels. At test time, the learned model predicts the stability results against the results generated by the simulator.

In a pilot study, we tested on a subset of the generated data with LeNet~\citep{lecun1995comparison}, a relatively small network designed for digit recognition, AlexNet~\citep{krizhevsky2012imagenet}, a large network and VGG Net~\citep{simonyan2014very}, an even larger network than AlexNet. We trained from scratch for the LeNet and fine-tuned for the large network pre-trained on ImageNet \cite{deng2009imagenet}. VGG Net consistently outperforms the other two, hence we use it across our experiment. We use the Caffe framework~\citep{jia2014caffe} in all our experiments.

We divide the experiment design into 3 sets: the intra-group, cross-group and generalization. The first set investigates influence on the model's performance from an individual scene parameter, the other two sets explore generalization properties under different settings.

%--------------------------------------------------------------------------------------------------------------------------------------
\subsubsection{Intra-Group Experiment}
In this set of experiments, we train and test on the scenes with the same scene parameters in order to assess the feasibility of our task.

\paragraph{Number of Blocks (4B, 6B, 10B, 14B)}
In this group of experiment, we fix the stacking depth and keep the all blocks in the same size but vary the number of blocks in the scene to observe how it affects the prediction rates from the image trained model, which approximates the relative recognition difficulty from this scene parameter alone. The results are shown in Table~\ref{tab:intra_group}. A consistent drop of performance can be observed with increasing number of blocks in the scene under various block sizes and stacking depth conditions. More blocks in the scene generally leads to higher scene structure and hence higher difficulty in perception. 

\paragraph{Block Size (Uni. vs. NonUni.)}
In this group of experiment, we aim to explore how same size and varied blocks sizes affect the prediction rates from the image trained model. We compare the results at different number of blocks to the previous group, in the most obvious case,  scenes happened to have similar stacking patterns and same number of blocks can result in changes visual appearance. To further eliminate the influence from the stacking depth, we fix all the scenes in this group to be 2D stacking only. As can be seen from Table~\ref{tab:intra_group}, the performance decreases when moving from 2D stacking to 3D. The additional variety introduced by the block size indeed makes the task more challenging.

\paragraph{Stacking Depth (2D vs. 3D)}
In this group of experiment, we investigate how stacking depth affects the prediction rates. With increasing stacking depth, it naturally introduces ambiguity in the perception of the scene structure, namely some parts of the scene can be occluded or partially occluded by other parts. Similar to the experiments in previous groups, we want to minimize the influences from other scene parameters, we fix the block size to be the same and only observe the performance across different number of blocks. The results in Table~\ref{tab:intra_group} show a little inconsistent behaviors between relative simple scenes (4 blocks and 6 blocks) and difficult scenes (10 blocks and 14 blocks). For simple scenes, prediction accuracy increases when moving from $2D$ stacking to $3D$ while it is the other way around for the complex scene. Naturally relaxing the constraint in stacking depth can introduce additional challenge for perception of depth information, yet given a fixed number of blocks in the scene, the condition change is also more likely to make the scene structure lower which reduces the difficulty in perception. A combination of these two factors decides the final difficulty of the task, for simple scenes, the height factor has stronger influence and hence exhibits better prediction accuracy for $3D$ over $2D$ stacking while for complex scenes, the stacking depth dominates the influence as the significant higher number of blocks can retain a reasonable height of the structure, hence receives decreased performance when moving from $2D$ stacking to $3D$.

\begin{table}
\centering
\ra{1.4}
\begin{tabular*}{0.7\linewidth}{@{ }ccccc@{ }}\toprule
\# Blks & \multicolumn{2}{c}{Uni.} & \phantom{ab}& \multicolumn{1}{c}{NonUni.}\\
\cmidrule{2-3} \cmidrule{5-5}
 & $2D$ & $3D$ && $2D$\\ \midrule
$4B$ & 93.0 & 99.2 && 93.2\\
$6B$ & 88.8& 91.6&& 88.0\\
$10B$ & 76.4& 68.4&& 69.8\\
$14B$ & 71.2& 57.0&& 74.8\\ \bottomrule
\end{tabular*}
\caption{Intra-group experiment by varying scene parameters.} 
\label{tab:intra_group}
\end{table}

%--------------------------------------------------------------------------------------------------------------------------------------
\subsubsection{Cross-Group Experiment}
In this set of experiment, we want to see how the learned model transfers across scenes with different complexity, so we further divide the scene groups into two large groups by the number of blocks, where a {\it simple scene} group for all the scenes with $4$ and $6$ blocks and a {\it complex scene} for the rest of scenes with $10$ and $14$ blocks. We investigate in two-direction classification, shown in the figure in Table~\ref{tab:cross_group}:
\begin{enumerate}
\item
Train on simple scenes and predict on complex scenes: Train on 4 and 6 blocks and test on 10 and 14 blocks
\item
Train on complex scenes and predict on simple scenes: Train on 10 and 14 blocks and test on 4 and 6 blocks
\end{enumerate}

As shown in Table~\ref{tab:cross_group}, when trained on simple scenes and predicting on complex scenes, it gets $69.9\%$, which is significantly better than random guess at $50\%$. This is understandable as the learned visual feature can transfer across different scene. Further we observe significant performance boost when trained on complex scenes and tested on simple scene. This can be explained by the richer feature learned from the complex scenes with better generalization. 

\begin{table}
\addtolength{\tabcolsep}{-3pt}
\centering
\ra{1.4}
\begin{tabular*}{0.95\linewidth}{l c c}
\multicolumn{3}{c}{\includegraphics[width=0.8\linewidth]{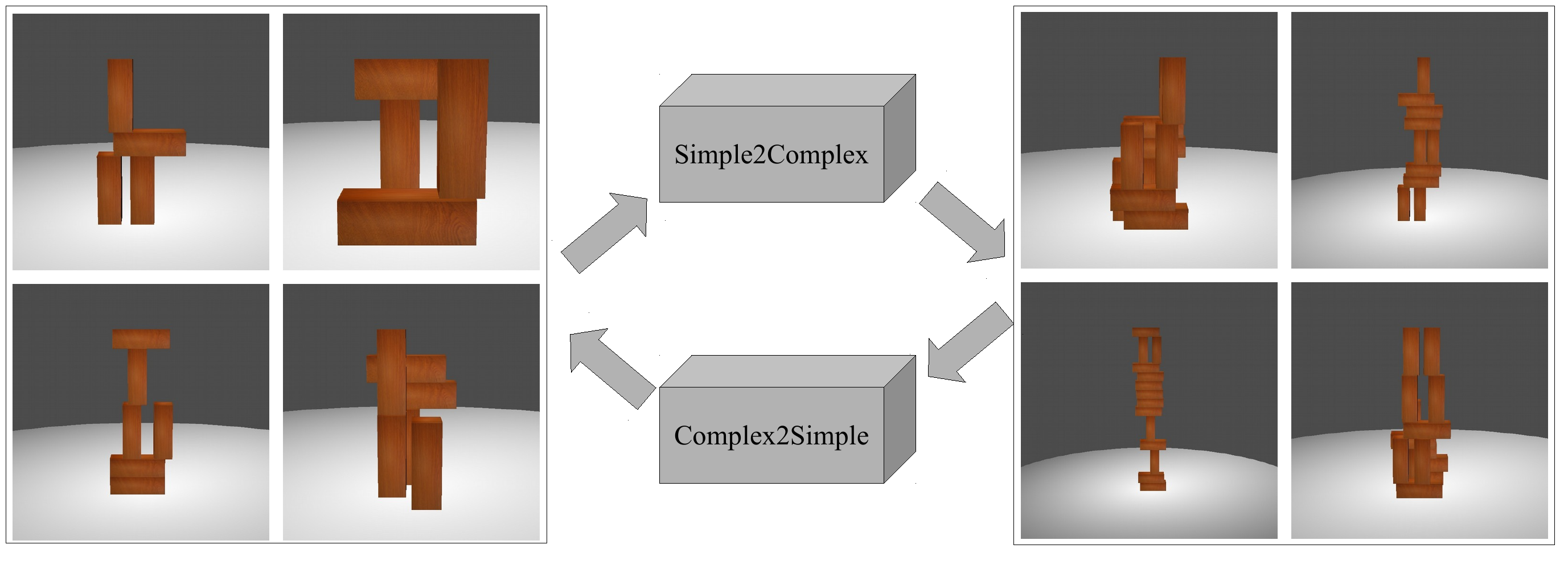}}\\
\toprule
Setting & \scriptsize Simple $\rightarrow$ Complex & \scriptsize Complex $\rightarrow$ Simple\\
\hline
Accuracy (\%)& 69.9 & 86.9\\
\bottomrule
\end{tabular*}
\addtolength{\tabcolsep}{3pt}
\caption{The upper figure shows the experiment settings for Cross-group classification where we train on simpler scenes and test on more complex scenes. The lower table shows the results.}
\label{tab:cross_group}
\end{table}

%--------------------------------------------------------------------------------------------------------------------------------------
\subsubsection{Generalization Experiment}
In this set of experiment, we want to explore if we can train a general model to predict stability for scenes with any scene parameters, which is very similar to human's prediction in the task. We use training images from all different scene groups and test on any groups. The Result is shown in Table~\ref{tab:general}. While the performance exhibits similar trend to the one in the intra-group with respect to the complexity of the scenes, namely increasing recognition rate for simpler settings and decreasing rate for more complex settings, there is a consistent improvement over the intra-group experiment for individual groups. Together with the result in the cross-group experiment, it suggests  a strong generalization capability of the image trained model.

\begin{table}
\centering
\ra{1.4}
\begin{tabular*}{0.75\linewidth}{@{ }cccccc@{ }}\toprule
\# Blks & \multicolumn{2}{c}{Uni.} & \phantom{ab}& \multicolumn{2}{c}{NonUni.}\\
\cmidrule{2-3} \cmidrule{5-6}
 & $2D$ & $3D$ && $2D$ & $3D$\\ 
 \midrule
4B  &93.2 &99.0 &&95.4 &99.8 \\
6B  &89.0 &94.8 &&87.8 &93.0 \\
10B &83.4 &76.0 &&77.2 &74.8 \\
14B &82.4 &67.2 &&78.4 &66.2 \\ 
\bottomrule
\end{tabular*}
\caption{Results for generalization experiments.} 
\label{tab:general}
\end{table}

\begin{table}
\addtolength{\tabcolsep}{-3pt}
\centering
\ra{1.3}
\begin{tabular*}{0.88\linewidth}{@{ }cccccc@{ }}\toprule
\# Blks & \multicolumn{2}{c}{Uni.} & \phantom{a}& \multicolumn{2}{c}{NonUni.}\\
\cmidrule{2-3} \cmidrule{5-6}
& $2D$ & $3D$ && $2D$ & $3D$\\ 
 \midrule
4B  &\tiny79.1/\tiny{\bf91.7} &\tiny93.8/\tiny{\bf 100.0} &&\tiny72.9/\tiny{\bf 93.8} &\tiny92.7/\tiny{\bf 100.0} \\
6B  &\tiny78.1/\tiny{\bf 91.7} &\tiny83.3/\tiny{\bf 93.8} &&\tiny71.9/\tiny{\bf 87.5} &\tiny89.6/\tiny{\bf 93.8} \\
10B &\tiny67.7/\tiny{\bf 87.5} &\tiny72.9/\tiny72.9 &&\tiny66.7/\tiny{\bf 72.9} &\tiny71.9/\tiny68.8 \\
14B &\tiny71.9/\tiny{\bf 79.2} &\tiny68.8/\tiny66.7 &&\tiny71.9/\tiny{\bf 81.3} &\tiny59.3/\tiny{\bf 60.4} \\ 
\bottomrule
\end{tabular*}
\addtolength{\tabcolsep}{3pt}
\caption{Results from human subject test $a$ and corresponded accuracies from image-based model $b$ in format $a/b$ for the sampled data.} 
\label{tab:human_test}
\end{table}

\subsubsection{Discussion}
Overall, we can conclude that direct stability prediction is possible and in fact fairly accurate at recognition rates over $80\%$ for moderate difficulty levels. As expected, the 3D setting adds difficulties to the prediction from appearance due to significant occlusion for towers of more than 10 blocks. Surprisingly, little effect was observed for small tower sizes switching from uniform to non-uniform blocks - although the appearance difference can be quite small.
To better understand our results, we further discuss the following two questions:

\begin{figure*}
\centering
	\begin{subfigure}[c]{0.6\linewidth}
		\centering
		\includegraphics[width=1.\linewidth]{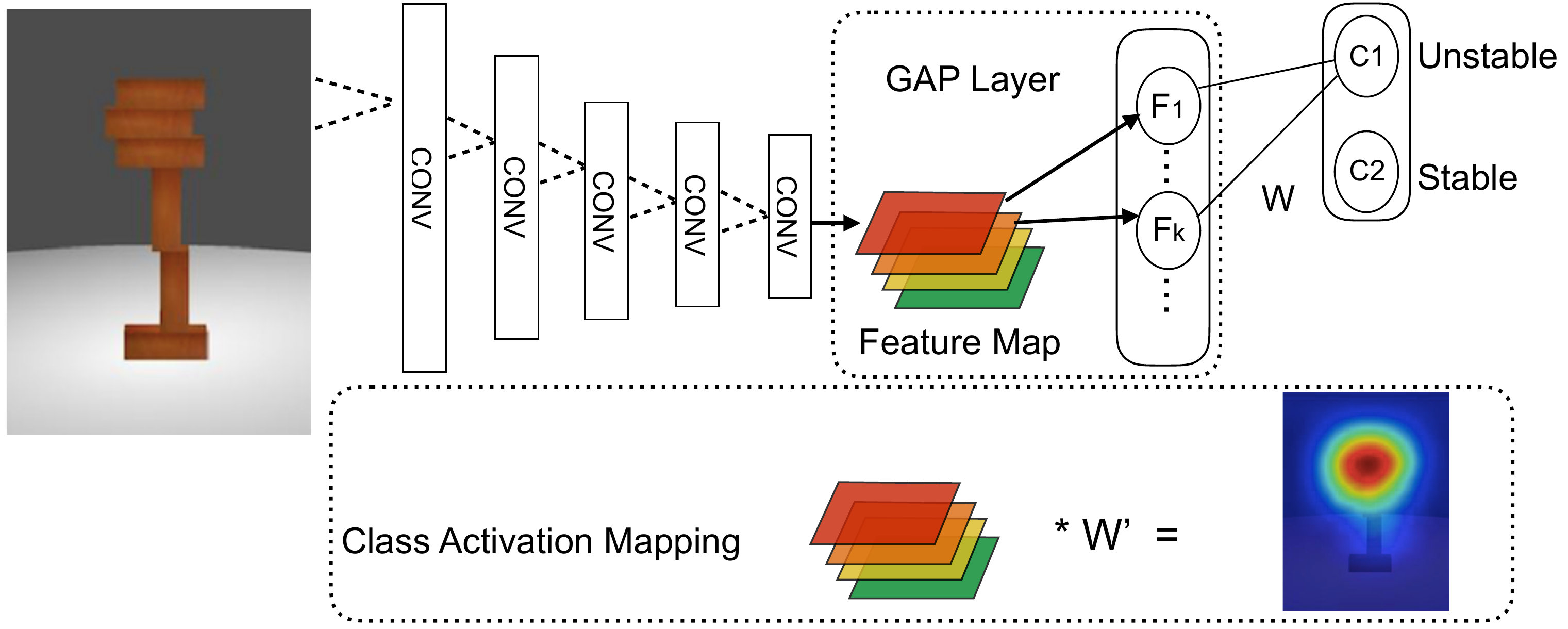}
		\caption{By introducing the GAP layer directly connected to the final output, the learned weights can be backprojected to the feature map for each category to construct the CAM. The CAM can be used to visualize the discriminative image regions for individual category.}
	\label{fig:gap_pipe}
	\end{subfigure}
	\begin{subfigure}[c]{0.34\linewidth}
		\centering
		\includegraphics[width=1.\linewidth]{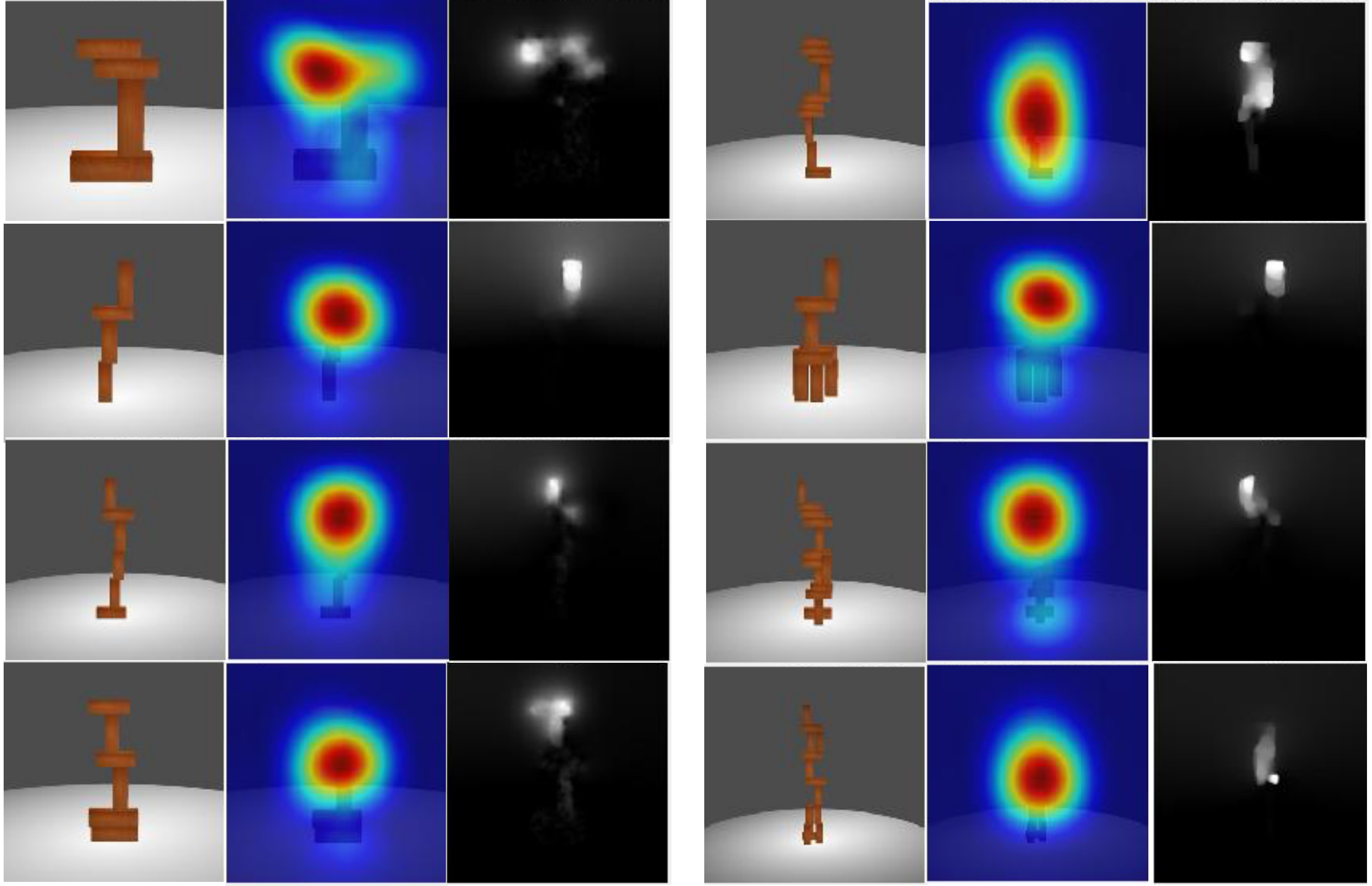}
		\caption{Examples of CAM showing the discriminative regions for unstable prediction in comparison to the flow magnitude indicating where the collapse motion begins. For each example, from left to right are original image, CAM and flow magnitude map.}
	\label{fig:gap_example}
	\end{subfigure}
\caption{We use CAM to visualize the results for model interpretation.}
\label{fig:gap}
\end{figure*}

\textbf{How does the model performs compared to human?} To answer this, we conduct a human subject test. We recruit human subjects to predict stability for give scene images. Due to large number of test data, we sample images from different scene groups for human subject test. 8 subjects are recruited for the test. Each subject is presented with a set of captured images from the test split. Each set includes $96$ images where images cover all $16$ scene groups with $6$ scene instances per group. For each scene image, subject is required to rate the stability on a scale from $1-5$ without any constraint for response time:
\begin{enumerate}
\item
Definitely unstable: definitely at least one block will move/fall
\item
Probably unstable: probably at least one block will move/fall
\item
Cannot tell: the subject is not sure about the stability
\item
Probably stable: probably no block will move/fall
\item
Definitely stable: definitely no block will move/fall
\end{enumerate}

The predictions are binarized, namely 1) and 2) are treated as unstable prediction, 4) and 5) as stable prediction, ``Cannot tell'' will be counted as $0.5$ correct prediction. 

The results are shown in Table~\ref{tab:human_test}. For simple scenes with few blocks, human can reach close to perfect performance while for complex scenes, the performance drops significantly to around $60\%$. Compared to  human prediction in the same test data, the image-based model outperforms human in most scene groups. While showing similar trends in performance with respect to different scene parameters, the image-based model is less affected by a more difficult scene parameter setting, for example, given the same block size and stacking depth condition, the prediction accuracy decreases more slowly than the counter part in human prediction. We interpret this as image-based model possesses better generalization capability than human in the very task. 

\textbf{ Does the model learn something explicitly interpretable?}  Here we apply the technique from \cite{zhou2015cnnlocalization} to visualize the learned discriminative image regions from CNN for individual category. The approach is illustrated in Figure~\ref{fig:gap_pipe}. With Global Average Pooling (GAP), the resulted spatial average of the feature maps from previous convolutional layers forms fully-connected layer to directly decides the final output. By back-projecting the weights from the fully-connected layer from each category, we can hence obtain Class Activation Map (CAM) to visualize the discriminative image regions. In our case, we investigate discriminative regions for unstable predictions to see if the model can spot the weakness in the structure. We use deep flow\cite{weinzaepfel2013deepflow} to compute the optical flow magnitude between the frame before the physics engine is enabled and the one afterwards to serve as a coarse ground truth for the structural weakness where we assume the collapse motion starts from such weakness in the structure. Though not universal among the unstable cases, we do find significant positive cases showing high correlation between the activation regions in CAM for unstable output and the regions where the collapse motion begins. Some examples are shown in Figure~\ref{fig:gap_example}.

\begin{figure}
\centering
\includegraphics[width=0.7\linewidth]{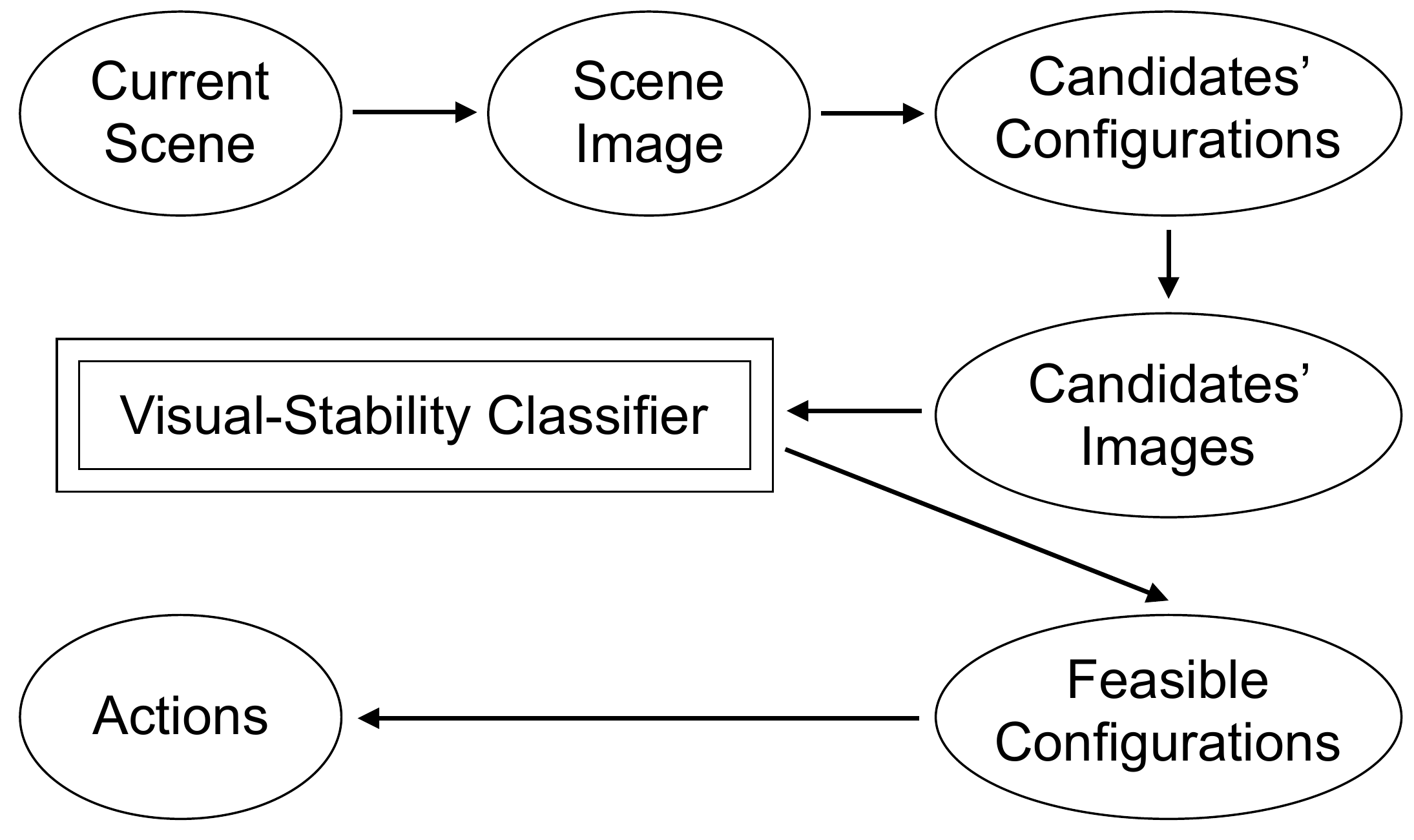}
\caption{An overview of our manipulation system. Our visual-stability classifier is integrated to recognize feasible candidate placement to guide manipulation.}
\label{fig:manip}
\end{figure}

\begin{figure}
\centering
\begin{subfigure}[b]{0.55\linewidth}\includegraphics[width=1\linewidth]{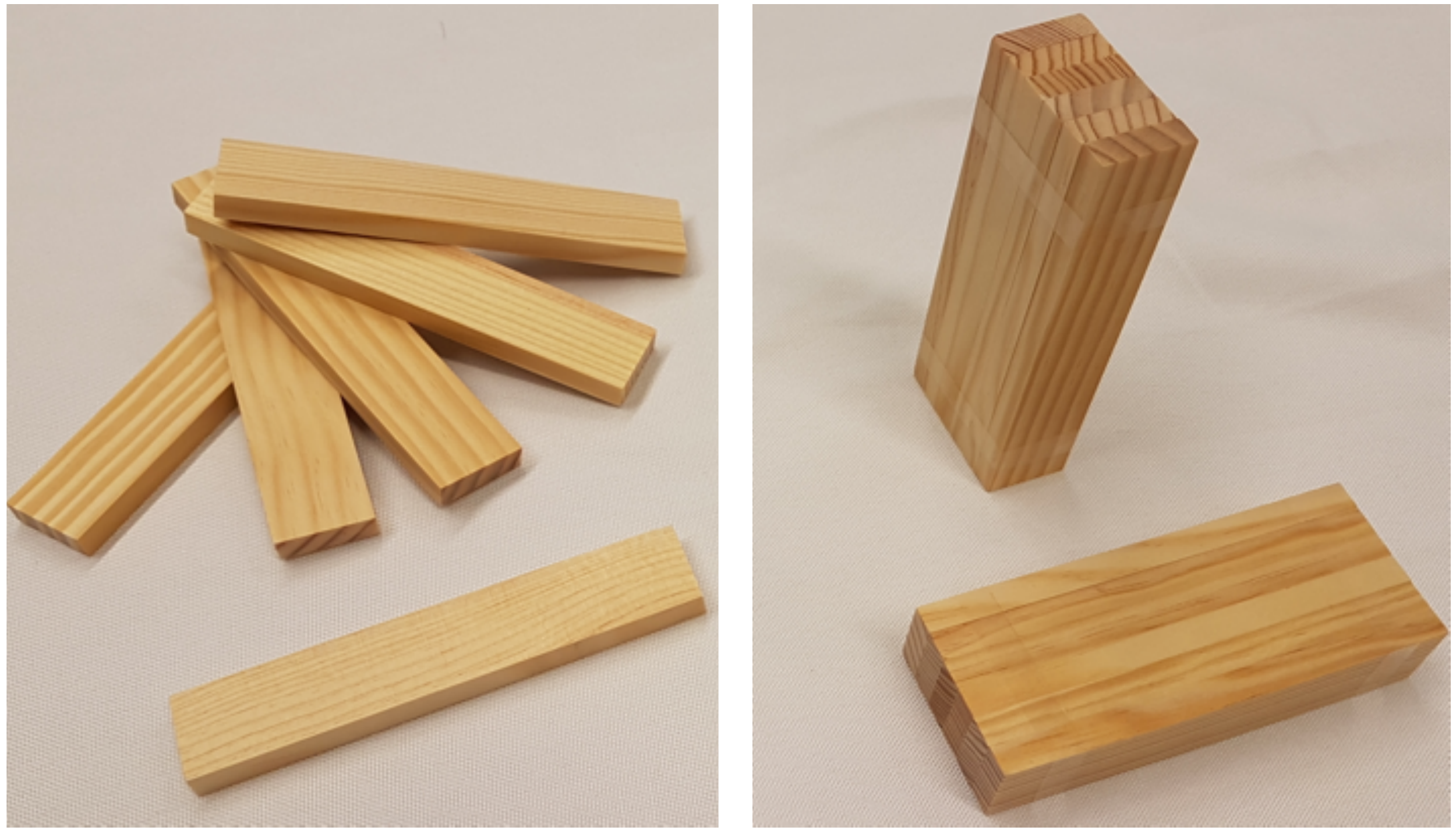}
\caption{Kapla block (left), block in test (right).}
\label{fig:blks}\end{subfigure}
\begin{subfigure}[b]{0.34\linewidth}\includegraphics[width=1\linewidth]{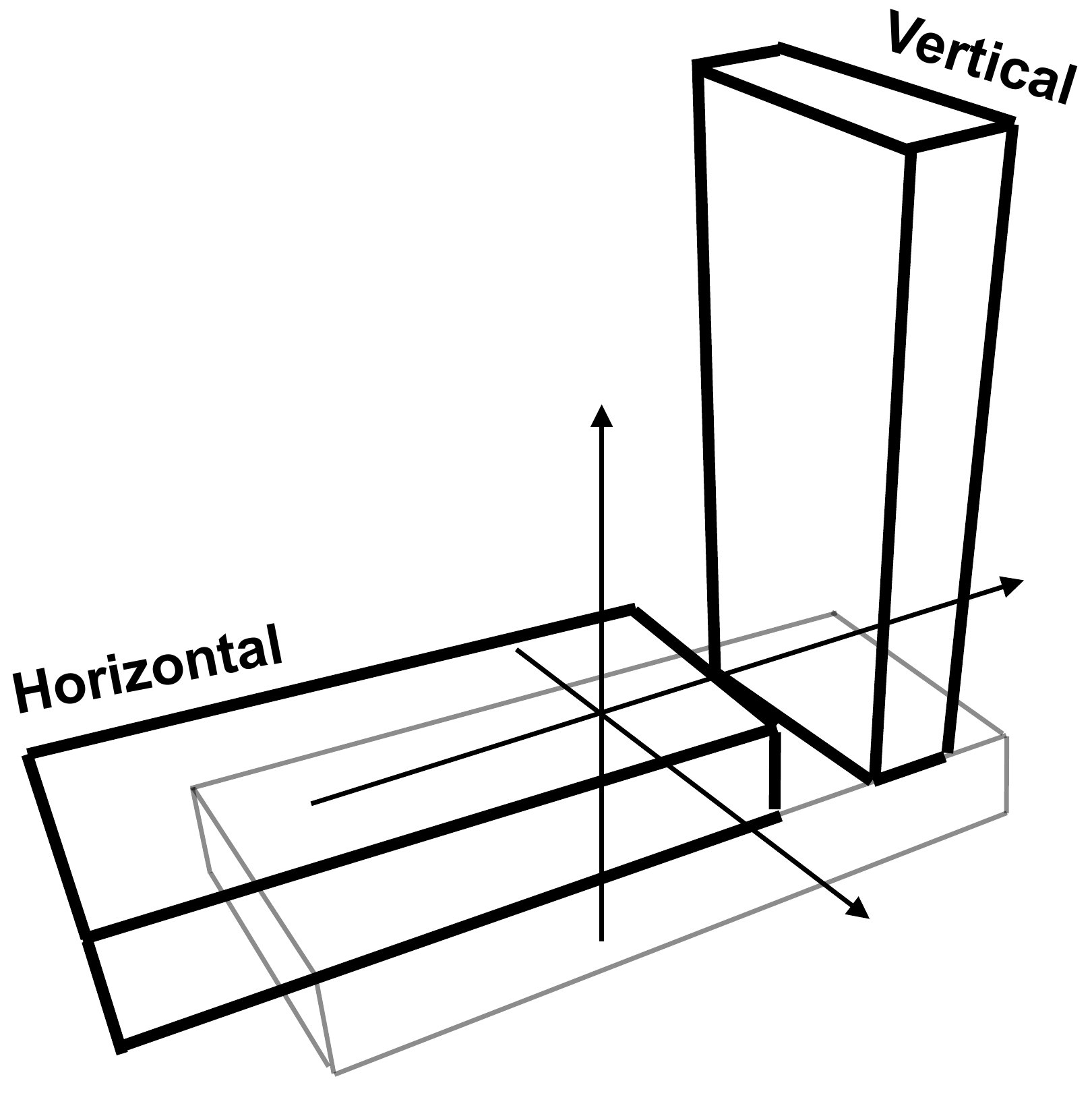}
\caption{Allowed configurations in test.}
\label{fig:blk_config}\end{subfigure}
\caption{Blocks used in our experiment.}
\label{fig:overview_manip}
\end{figure}

In the previous section, we have shown that an appereance-based model can predict  physical stability relatively well on the synthetic data. Now we want to further explore if and how the synthetic data trained model can be utilized for a real world application, especially for robotic manipulation. 
Hence, we decide to set up a testbed where a Baxter robot's task is to stack one wood block on a given block structure without breaking the structure's stability as shown in Figure~\ref{fig:teaser}. The overview of our system is illustrated in Figure~\ref{fig:manip}. In our experiment, we use Kapla blocks as basic unit, and tape 6 blocks into a bigger one as shown in Figure~\ref{fig:blks}. To simplify the task, adjustments were made to the free-style stacking:

\begin{itemize}
\item The given block structure is restricted to be single layer as the ${\text 2D}$ case in the previous section. For the final test, we report results on the 6 scenes as shown in Table~\ref{tab:real_task}.
\item The block to be put on top of the given structure is limited two canonical configurations \{vertical, horizontal\} as shown in Figure~\ref{fig:blk_config}. and assumed to be held in hand of robot before the placement.
\item The block is constrained to be placed on the top-most horizontal surface (stacking surface) in the given structure.
\item The depth of the structure (perpendicular distance to the robot) is calibrated so that we only need to decide the horizontal and vertical displacements with respect to the stacking surface. 
\end{itemize}

\subsection{Prediction on Real World Data}

Considering there are significant difference between the synthesized data and real world captured data, including factors (not limited to) as texture, illumination condition, size of blocks and accuracy of the render, we performed a pilot study to directly apply the model trained on the RGB images to predict stability on the real data, but only got results on par with random guessing. Hence we decided to train the visual-stability model on the binary-valued foreground mask on the synthesized data and deal with the masks at test time also for the real scenes. In this way, we significantly reduce the effect from the aforementioned factors. Observing comparable results when using the RGB images, we continue to the approach on real world data.

\begin{figure}
\centering
\includegraphics[width=1\linewidth]{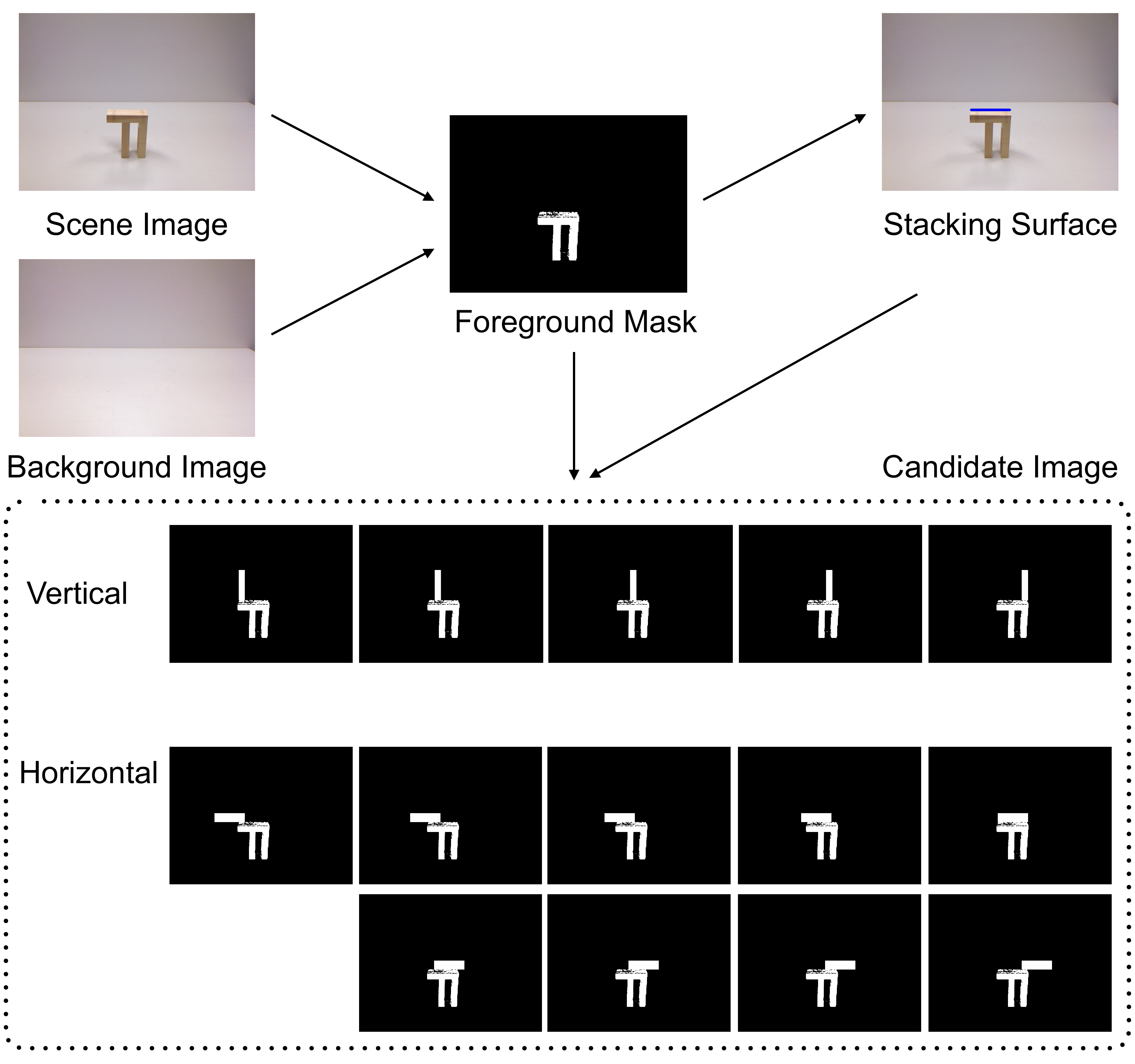}
\caption{The procedure to generate candidates placement images for a give scene in our experiment.}
\label{fig:candidates}
\end{figure}

At test time, a background image is first captured for the empty scene. Then for each test scene (shown in Table~\ref{tab:real_task}), an image is captured and converted to foreground mask via background subtraction. The top-most horizontal boundary is detected as the stacking surface and then used to generate candidate placements: the surface is divided evenly into 9 horizontal candidates and 5 vertical candidates, resulting in 84 candidates. The process is shown in Figure~\ref{fig:candidates}. Afterwards, these candidates are put to the visual-stability model for stability prediction. Each generated candidate's actual stability is manually tested and recorded as ground truth. The final recognition result is shown in Table~\ref{tab:real_task}. The model trained with synthetic data is able to predict with overall accuracy of $78.6\%$ across different candidates in real world.

\subsection{Manipulation Test}
At test time, when the model predicts a give candidate placement as stable, the robot will execute routine to place the block with 3 attempts. We count the execution as a success if any of the attempt works. The manipulation success rate is defined as:
\begin{gather*}
\begin{aligned}
\frac{\text{\#\{successful placements\}}}{\text{\#\{all stable placements\}}}
\end{aligned}
\end{gather*}
where $\text{\#\{successful placements\}}$ is the number of successful placements made by the robot, and $\text{\#\{all stable placements\}}$ is the number of all ground truth stable placements.

\begin{table*}
\addtolength{\tabcolsep}{-4.3pt}
\centering
\ra{1.4}
\begin{tabular*}{1.\linewidth}{crrrrrrrrrrrr}%\toprule
\hline
Id. & \multicolumn{2}{c}{1} & \multicolumn{2}{c}{2} & \multicolumn{2}{c}{3} & \multicolumn{2}{c}{4} &\multicolumn{2}{c}{5} &\multicolumn{2}{c}{6}\\
\hline 
%&&&&&&\\
Scene & 
\multicolumn{2}{c}{\includegraphics[width=0.1\linewidth]{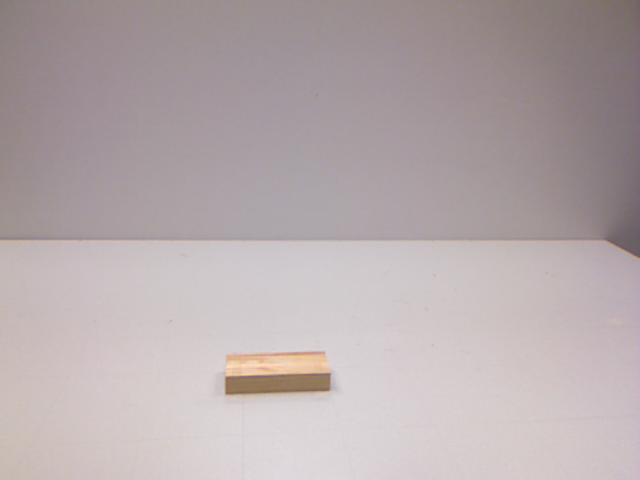}} &
\multicolumn{2}{c}{\includegraphics[width=0.1\linewidth]{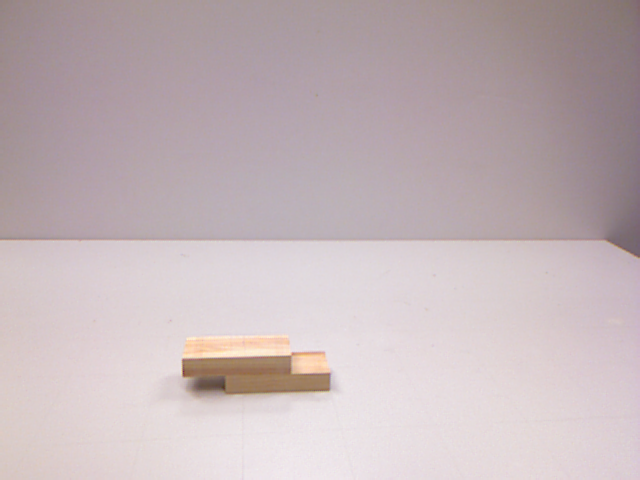}} &
\multicolumn{2}{c}{\includegraphics[width=0.1\linewidth]{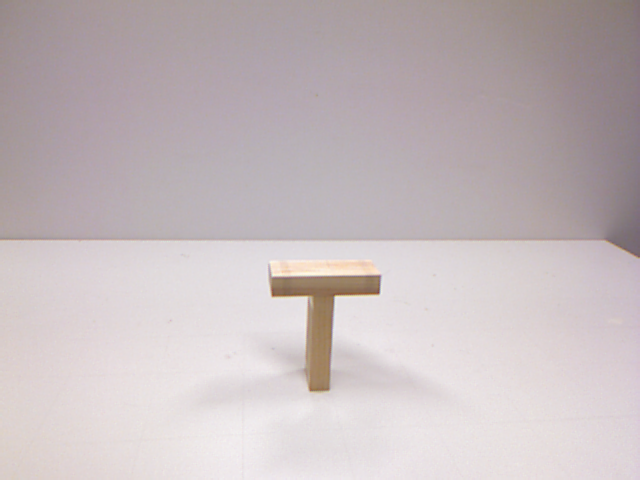}} &
\multicolumn{2}{c}{\includegraphics[width=0.1\linewidth]{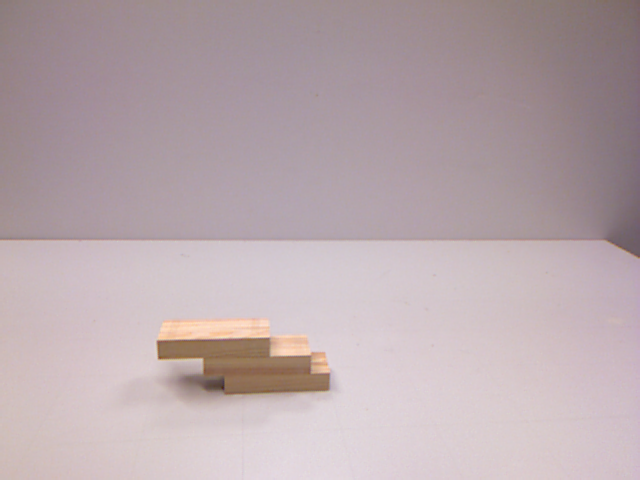}} &
\multicolumn{2}{c}{\includegraphics[width=0.1\linewidth]{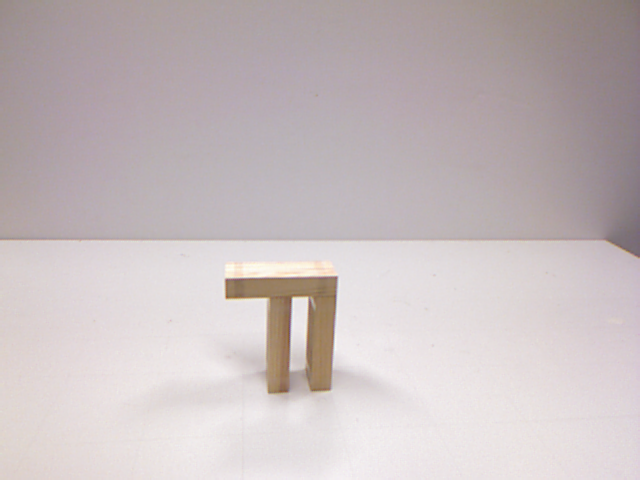}} &
\multicolumn{2}{c}{\includegraphics[width=0.1\linewidth]{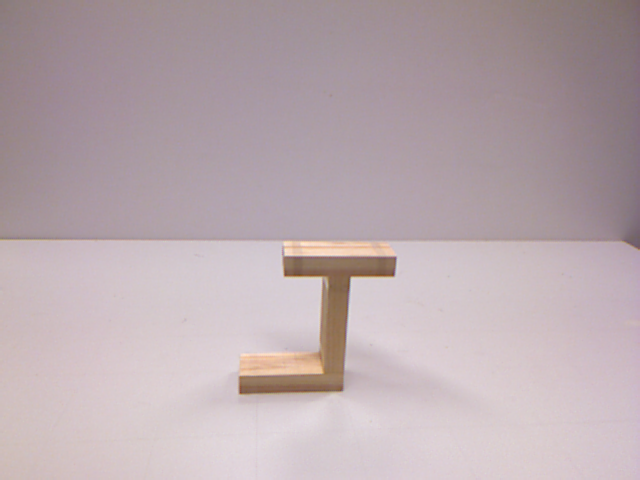}} \\
\hline
Pred.($\%$)& 

\footnotesize66.7& \footnotesize100.0& 
\footnotesize66.7& \footnotesize60.0& 
\footnotesize88.9& \footnotesize100.0& 
\footnotesize77.8& \footnotesize80.0& 
\footnotesize100.0& \footnotesize40.0& 
\footnotesize66.7& \footnotesize60.0\\
Mani.($\%$)&
\footnotesize80.0\tiny(4/5)& \footnotesize100.0\tiny(5/5)&
\footnotesize66.7\tiny(2/3)& \footnotesize100.0\tiny(3/3)&
\footnotesize66.7\tiny(2/3)& \footnotesize100.0\tiny(1/1)&
\footnotesize66.7\tiny(2/2)& \footnotesize66.7\tiny(2/3)&
\footnotesize100.0\tiny(3/3)& \footnotesize25.0\tiny(1/4)&
\footnotesize0.0\tiny(0/3)& \footnotesize0.0\tiny(0/1)\\
\bottomrule
Placement& 
H& V&
H& V&
H& V&
H& V&
H& V&
H& V\\
\hline
\end{tabular*}
\addtolength{\tabcolsep}{6pt}
\caption{Results for real world test. ``Pred.'' is the prediction accuracy. ``Mani.'' is the manipulation success rate with counts for successful placements/all possible stable placements for each scene. ``H/V'' refer to horizontal/vertical placement.} 
\label{tab:real_task}
\end{table*}

As shown in Table~\ref{tab:real_task}, the manipulation performance is good across most of the scenes for both horizontal and vertical placements except for the 6-th scene where the classifier predicts all candidates as unstable hence no attempts have been made by the robot.

\subsection{Discussion}
Simply putting the block along the center of mass (COM) of the given structure may often be a feasible option, yet, there are two limitations to this approach: first, it is nontrivial to compute the COM of a given structure; second, it only gives one possible stable solution (assuming it actually stay stable). In comparison, our method does not rely the COM of the structure and provide a search over multiple possible solutions. 

So far, we have successfully applied our visual stability prediction model to guide the stable placement of a single block onto the existing structure. However, this approach has its own limitations. Firstly, it only deals with one manipulation step and does not take into account the interaction process which is also under physics constraint. For instance, the block can collide with existing structure before its placement. Secondly, it is also not obvious how to extend the approach to plan a consecutive stacking of multiple blocks. In part II, we will explore an alternative approach to overcome these limitations.
\section{Part II: Beyond Single Block Stacking --- Target Stacking}
In this part, we look at a manipulation task beyond the single block placement as discussed in part I. We now seek solutions to plan the placement of multiple blocks and consider more complex interactions in physical environment. In particular, just like children stack blocks towards a certain blueprint, we further introduce the "target" into the stacking task so that the sequence of manipulations are planned to a achieve the target --- similar to planning. Yet we want to stay fully in an end-to-end learning approach -- concerning the perception, physics as well as the plan, where intuitive physics as well as plan execution is learned jointly.

The extended block stacking task is called {\em target stacking\/}. In this task, an image of a target structure made of stacked blocks is provided. Given the same number of blocks as in the target structure, the goal is to reproduce the structure shown in the image. The manipulation primitives in this task include moving and placing blocks. This is inspired by the scenario where young children learn to stack blocks to different shapes given an example structure. We want to explore how an artificial agent can acquire such a skill through trial and error. 

\subsection{Task Description}
\begin{figure}
\centering
\includegraphics[width=0.95\linewidth]{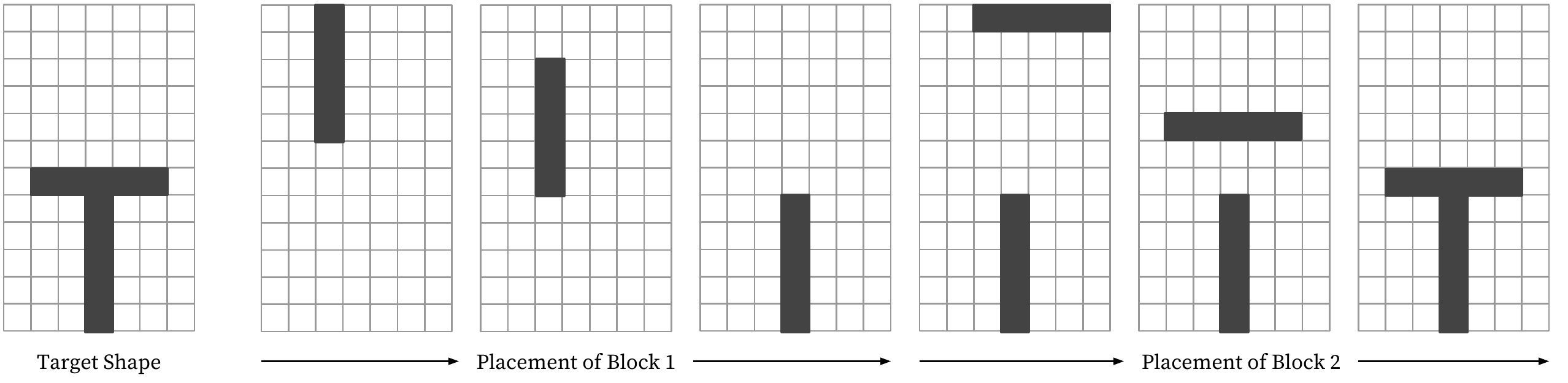}
\caption{Target stacking: Given a target shape image, the agent is required to move and stack blocks to reproduce it.}
\label{fig:tstack_demo}
\end{figure}

For each task instance, a target structure is generated and its image is provided to the agent along with the number of blocks. Each of these blocks has a fixed orientation. The sequence of block orientations is such that reproducing the target is feasible. The agent attempts to construct the target structure by placing the blocks in the given sequence. The spawning location for each block is randomized along the top boundary of the environment. A illustrative sample for the task is shown in Figure~\ref{fig:tstack_demo}.

\subsection{Task Distinction}
The following characteristics distinguish this task from other tasks commonly used in the literature.

\paragraph{Goal-Specific}
A widely-used benchmark for deep reinforcement learning algorithm are the Atari games \citep{bellemare13arcade} that were made popular by \cite{mnih2013playing}. While this game collection has a large variety, the games are defined by a single goal or no specific goal is enforced at a particular point in time. For example in Breakout, the player tries to bounce off as many bricks as possible. In Enduro, the player tries to pass as many cars as possible while simultaneously avoiding cars. 

In the target stacking task, each task instance differs in the specific goal (the target structure), and all the moves are planned towards this goal. Given the same state, moves that were optimal in one task instance are unlikely to be optimal in another task instance with a different target structure. This is in contrast to games where one type of move will most likely work in similar scenes. This argument also applies to AI research platforms with richer visuals like VizDoom \citep{kempka2016vizdoom}.

\paragraph{Longer sequences}
Target stacking requires looking ahead over a longer time horizon to simultaneously ensure stability and similarity to the target structure. This is different from learning to poke \citep{agrawal2016learning} where the objective is to select a motion primitive that is the optimal next action. It is also different from the work by \cite{li2017visual} that reasons about the placement of one block. 

\paragraph{Rich Physics Bounded}
Besides stacking to the assigned target shape the agent needs to learn to move the block without colliding with the environment and existing structure and to choose the block's placement wisely not to collapse the current structure. The agent has no prior knowledge of this. It needs to learn everything from scratch by observing the consequence (collision, collapse) of its actions.  

\subsection{Target Stacking Environment}
Compared to the environment for visual stability prediction as a data generator, we now need more a more capable task environment supports the learning process with which the agent can interact. This requires us to significantly extend the implementation. 

The environment now implements a more dynamic system where different moves from the agent can lead to different consequences: when a block (not the last one for the episode) is placed stably in the scene, a new block should be spawned into the environment whereas if the block collapses the existing structure, the current episode is terminated so that a new one can be started. The key to react correctly under these different conditions is to detect them effectively. 

While we keep the essential parts of the task, at its current stage the simulated environment remains an abstraction of a real-world robotics scenario. This generally requires an integration of multiple modules for a full-fledged working system, such as \cite{toussaint2010integrated}, which is out of scope of this paper.

\begin{figure}
\centering
\includegraphics[width=0.95\linewidth,keepaspectratio]{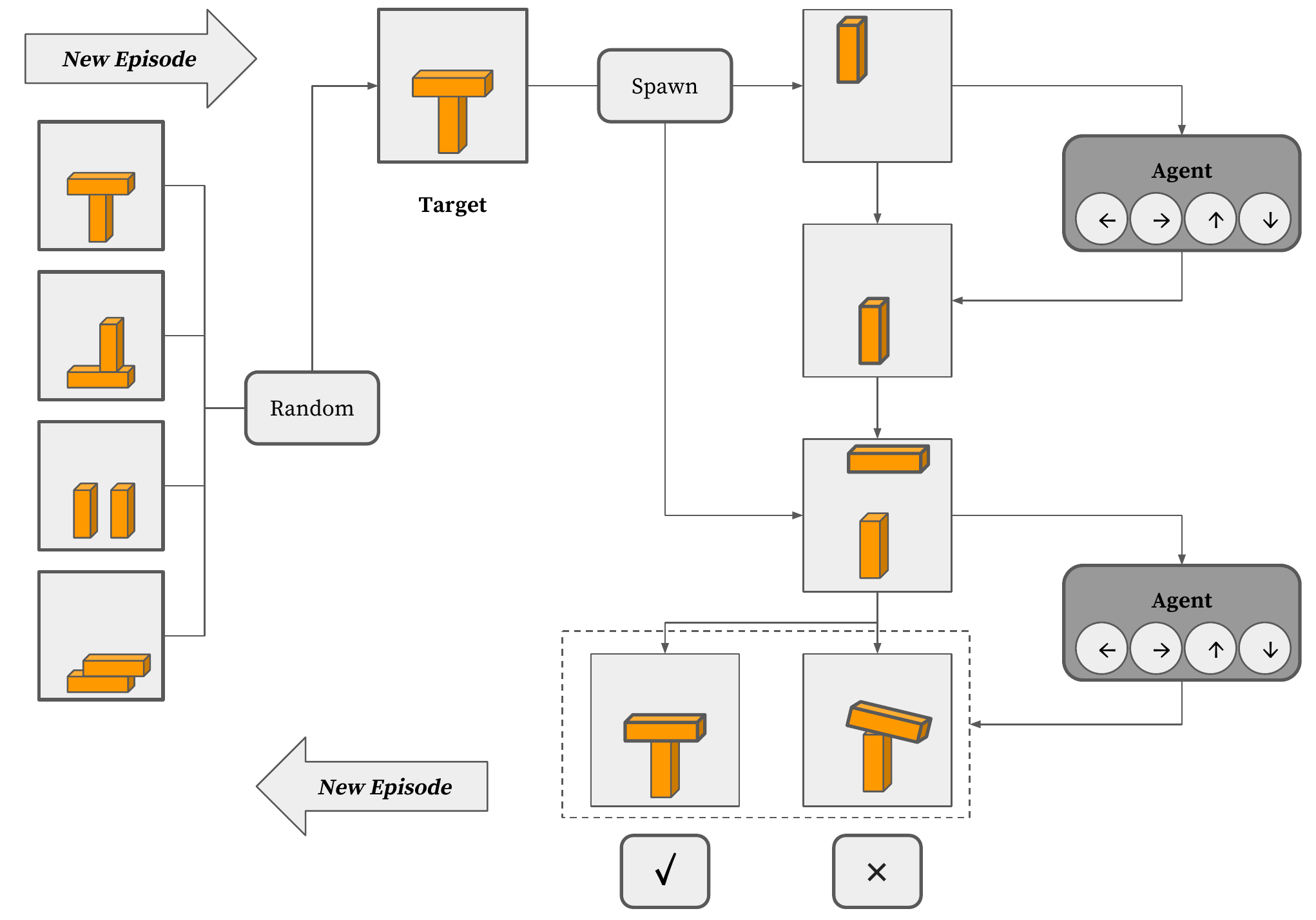}
\caption{Overview of design for the target stacking environment.}
\label{fig:sim:targetStack_pipeline}
\end{figure}

An overview of the stacking environment is shown in Figure~\ref{fig:sim:targetStack_pipeline}. The environment is again implemented in Panda3D as the data generator described in the previous part. The block size follows a ratio of $l:w:h = 5:2:1$, where $l$,$w$,$h$ denote length, width and height respectively.

For each episode, a target structure is randomly picked from a pool of structures and then the blocks are spawned according to the target. The agent moves the spawned block in the scene until the placement is done. 

During this process, if the agent's move causes the block to (1) move out over the boundary of the scene or (2) collide with existing structure (3) collapse the structure with the placement, the episode is terminated and a new episode can start. 

A new block is spawned upon the previous one is placed successfully till reaching the total number of blocks in the target. When all the blocks are placed in the scene, the environment determines if the agent achieves the assigned target structure. 

We ignore the impact during block placement and focus on the resulting stability of the entire structure. Once the block makes contact with the existing structure, it is treated as releasing the block for a placement. The physics simulation runs at $60Hz$. However considering the cost of simulation we only use it when there is contact between the moving block and the boundary or the existing structure. Otherwise, the current block is moving without actual physics simulation. 

\subsubsection{State Representation}
\begin{figure}
\centering
	\begin{subfigure}[b]{0.45\linewidth}
		\centering
		\includegraphics[width=1.\linewidth]{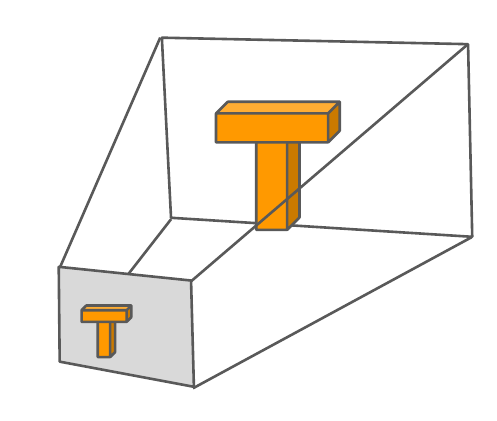}
		\caption{}
	\label{fig:sim:panda3d_perspective}
	\end{subfigure}
	\begin{subfigure}[b]{0.5\linewidth}
		\centering
		\includegraphics[width=1.\linewidth]{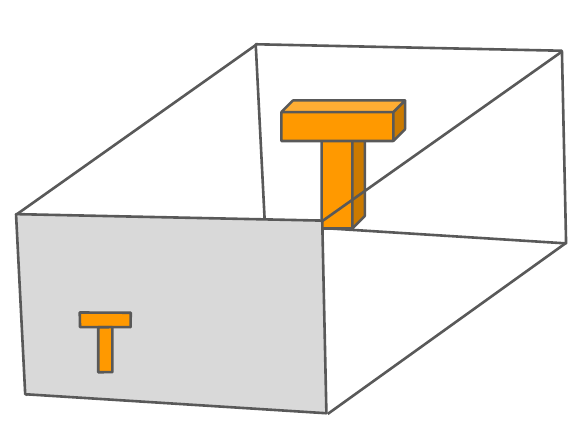}
		\caption{}
	\label{fig:sim:panda3d_orthographic}
	\end{subfigure}
\caption{Different perspective lens used in this thesis.(\protect\subref{fig:sim:panda3d_perspective}): Perspective projection. (\protect\subref{fig:sim:panda3d_orthographic}): Orthographic projection.}
\label{fig:sim:lense}
\end{figure}

To further reduce the appearance difference caused by varying perspective, the state in the stacking environment is rendered using orthographic projection. It is implemented by the \texttt{OrthographicLens} in Panda3D where parallel lines stay parallel and don't converge as shown in Figure~\ref{fig:sim:panda3d_orthographic}. This is in contrast with regular perspective camera used in the visual stability prediction environment as shown in Figure~\ref{fig:sim:panda3d_perspective}.

\begin{figure}[h]
\centering
\includegraphics[width=1.\linewidth,keepaspectratio]{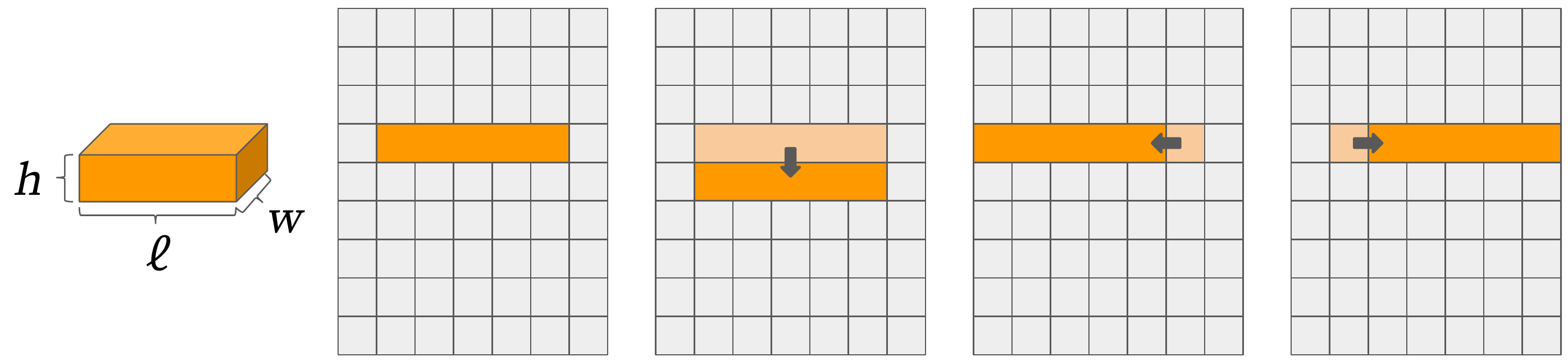}
\caption{State representation in the stacking environment}
\label{fig:sim:targetStack_state}
\end{figure}

Additionally, as we consider the action in discrete space, namely \{left,right,down\} for the task, the state representation can also be formalized correspondingly as shown in Figure~\ref{fig:sim:targetStack_state}. The individual block size follows a ratio of $l:w:h = 5:2:1$, where $l$,$w$,$h$ denote length, width and height respectively. With the aforementioned orthographic camera, only length and height are preserved on the projected image with a ratio of $l:h=5:1$. Each action moves the block by a displacement of the block's height along its direction. Hence the state can be represented in a grid map of unit square cells with length of the block's height ($u$ in the image space). In actual implementation, the blocks have to be set with coordinates in continuous values, so a separate binary matrix is used to maintain the discrete representation of the state where $1$ denotes the block presence in the cell and $0$ otherwise. 

The environment can query both the rendered image from the camera and the grid map for the state's representation, though in our current experiments, we use mostly the grid map for a simplified representation. Another benefit from the grid map representation is that it can be used as a occupancy map which is crucial for our implemented collision detection mechanism discussed in the next subsection. 

\subsubsection{Collision and Stability Detection}
As the block is maneuvered in the environment, it will eventually make contact with the scene. The occurrence of contact marks an important transition of the state for the environment. As shown in Figure~\ref{fig:sim:targetStack_detection}: 

\begin{figure}[h]
\centering
\includegraphics[width=.9\linewidth,keepaspectratio]{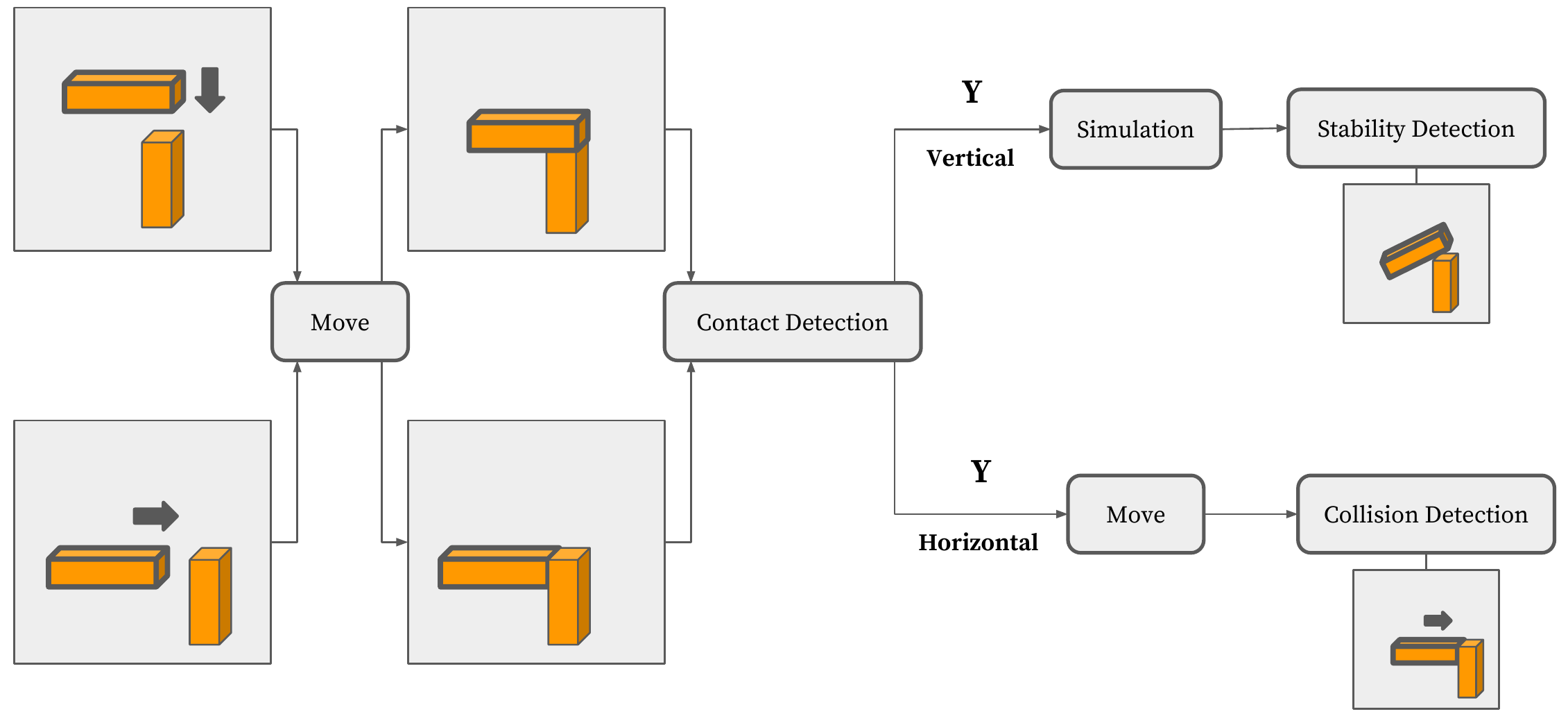}
\caption{Process of collision and stability detection}
\label{fig:sim:targetStack_detection}
\end{figure}

If the contact is made by a block from above (vertical), it suggests a placement \footnote{In our environment, we simplify the release movement from the real world block placement, i.e. the block is deemed to be released once it makes vertical contact with the scene, either the ground or the structure in the scene.}, then physics simulation is called to obtain the stability. The stability detection is implemented the same way as the one used in the data generator for visual stability prediction. Here the simulation only activates upon vertical contact is detected to save runtime.
If the block remains stable in the scene and is not the last block for the episode, a new block is then spawned at the top of the scene with random horizontal location. If the block is the final one and remains stable or it collapses the structure in the scene, then the current episode is terminated and a new episode can be started.

If the contact is made by a horizontal movement, it will not directly affect the current episode. If an additional move causes the collision, then the episode will be terminated, otherwise the block can be moved further until either vertical contact happens, it goes into the simulation and stability-detection branch or collision and terminated.

Note if the block is about to move out of the view of the camera, it is also considered a sub-type of collision. This is implemented straightforwardly by comparing the block's incoming location with the boundary of the scene. 

\begin{figure}
\centering
	\begin{subfigure}{0.98\linewidth}
		\centering
		\includegraphics[width=1.\linewidth]{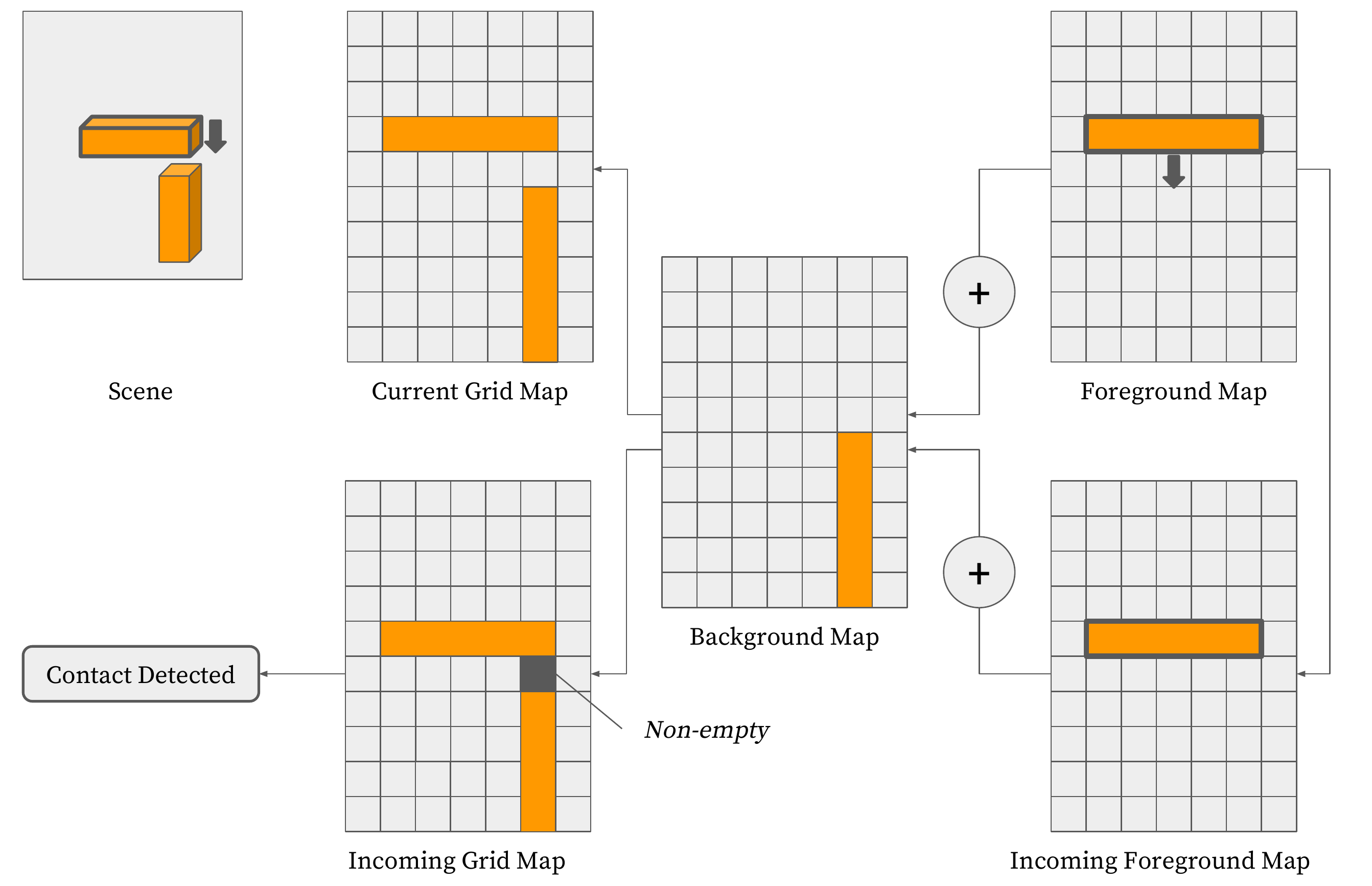}
		\caption{}
	\label{fig:sim:targetStack_contactDetect}
	\end{subfigure}
	\begin{subfigure}{0.98\linewidth}
		\centering
		\includegraphics[width=1.\linewidth]{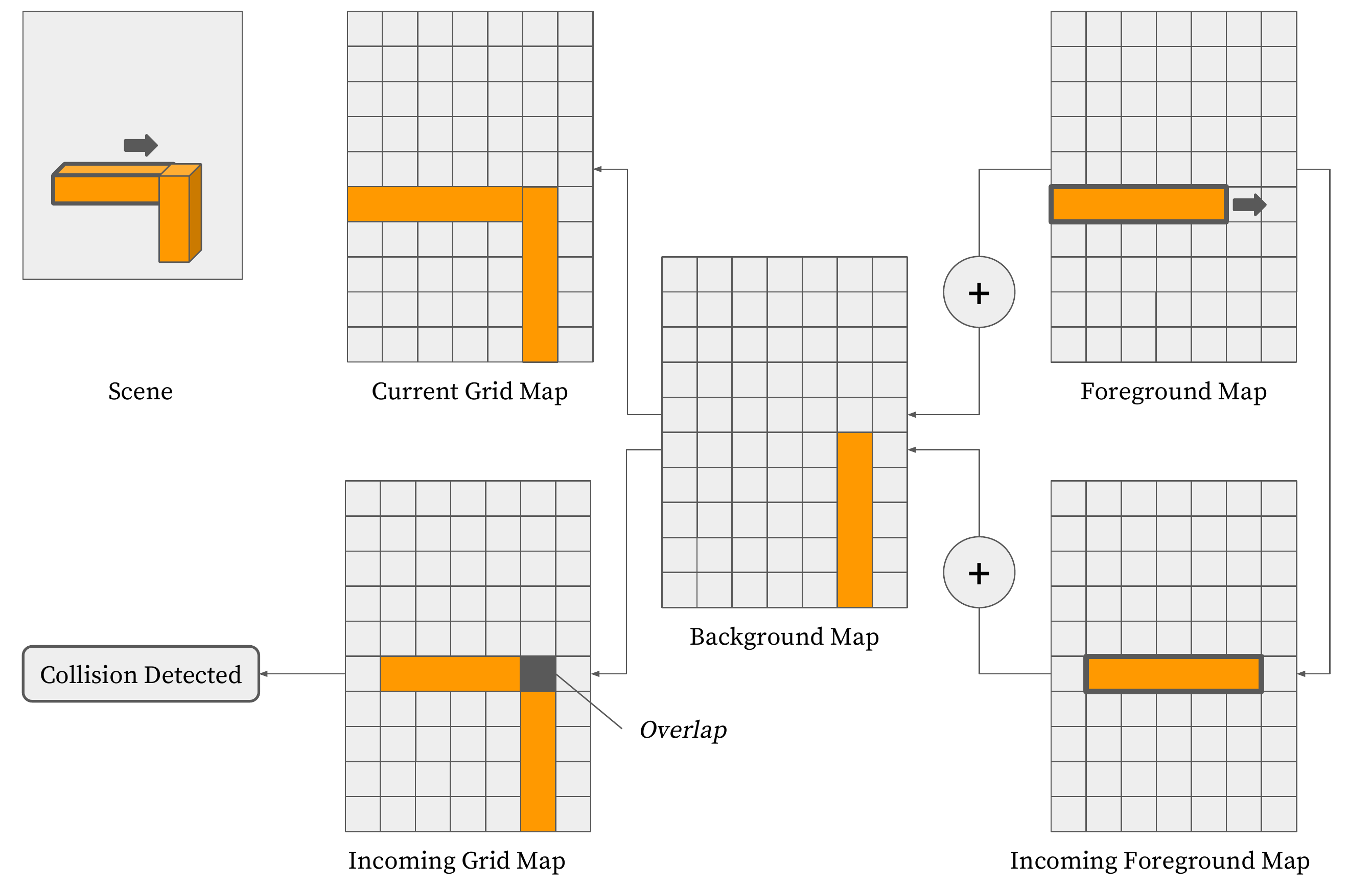}
		\caption{}
	\label{fig:sim:targetStack_collisionDetect}
	\end{subfigure}
\caption{Internal representation to detect contact and collision conditions: (\protect\subref{fig:sim:targetStack_contactDetect}) Detect vertical contact. (\protect\subref{fig:sim:targetStack_collisionDetect}) Detect collision.}
\label{fig:sim:targetStack_detection_detail}
\end{figure}

The contact and collision detection are implemented with the grid map representation described in the previous subsection as shown in Figure~\ref{fig:sim:targetStack_detection_detail}. We keep two grid map for each episode, one for the existing structure in the scene (\textbf{background map}) and the other for the moving block (\textbf{foreground map}). A state (grid map) is internally represented as the summation of the background map and foreground map. Given a state and a specified action, the system will compute the incoming foreground map after the action, if the incoming foreground map overlaps with the background map, then collision is detected as shown in Figure~\ref{fig:sim:targetStack_collisionDetect}; if any of the adjacent cells below the block in the incoming foreground map is non-empty in the incoming grid map, then a vertical contact is detected as shown in Figure~\ref{fig:sim:targetStack_contactDetect}.

\subsubsection{Interface Design}
\begin{figure}
\centering
\includegraphics[width=.98\linewidth,keepaspectratio]{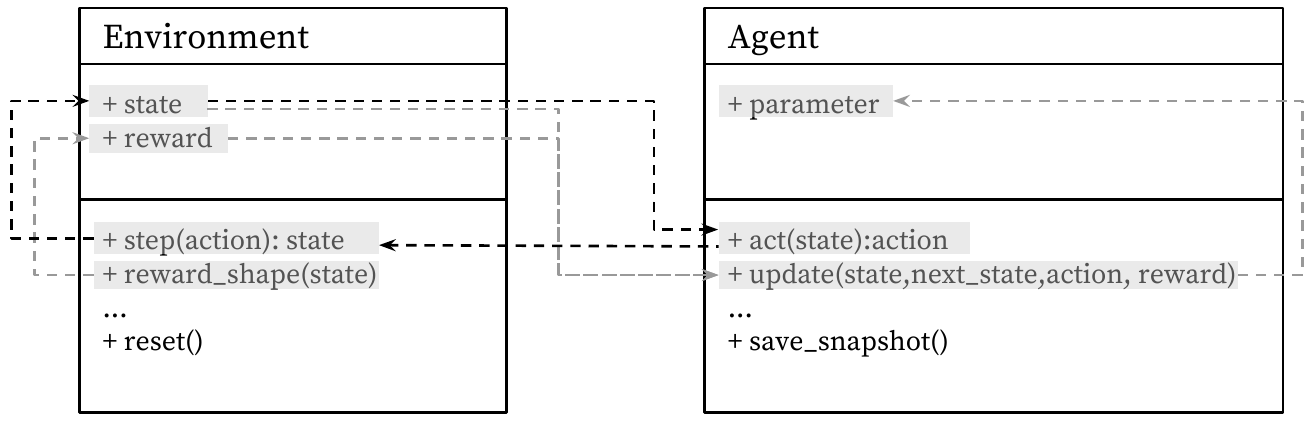}
\caption{Interface design for the environment with interaction of the agent.}
\label{fig:sim:targetStack_oop}
\end{figure}

With recent progress in reinforcement learning, there is an increasing need for the standardized benchmarks for different learning agents. To this end, OpenAI recently introduced Gym \citep{brockman2016openai} to provide access to a standardized set of environments, such as Atari games and board games. By encapsulating the environment's internal mechanism from the agent, different agents can be evaluated on the environment with minimal modification. 

We also adopt a similar design in our environment. The diagram in Figure~\ref{fig:sim:targetStack_oop} shows part of the design of the environment with interactions with an agent. The agent's \texttt{act} method takes the current state from the environment, decides its action and pass the decision to the environment. The \texttt{step} method receives the action from the agent and updates the environment's state. The \texttt{reward\_shape} method evaluates the current state and transforms the reward with specific designs. Then the pair of states before and after the action, together with the reward, are input to the \texttt{update} method of the agent to update its parameter depending on different learning algorithms. 

\subsection{Goal-Parameterized Deep Q Networks (GDQN)}
As one major characteristic of this task is that it requires goal-specific planning: given the same or similar states under different objectives, the optimal move can be different. To this end, we extend the typical reinforcement learning formulation to incorporate additional goal information.  

\subsubsection{Learning Framework}
In a typical reinforcement learning setting, the agent interacts with the environment at time $t$, observes the state $s_t$, takes action $a_t$, receives reward $r_t$ and transits to a new state $s_{t+1}$. A common goal for a reinforcement learning agent is to maximize the cumulative reward. This is commonly formalized in form of a value function as the expected sum of rewards from a state $s$, $\mathbf{E}[\sum\limits_{i=0}^T \gamma^i r_{t+i+1}|s_t = s,\pi]$ when actions are taken with respect to a policy $\pi(a|s)$, with $0\leq\gamma\leq 1$ being the discount factor, T for the final time step. The alternative formulation to this is the action-value function $Q^\pi(s,a)=\mathbf{E}[\sum\limits_{i=0}^T \gamma^i r_{t+i+1}|s_t=s,a_t=a]$. 

Value-based reinforcement learning algorithms, such as Q-learning \citep{watkins1992q} directly search for optimal Q-value function. Recently by incorporating deep neural network as a function approximator for $Q$-function, the DQN \citep{mnih2015human} has shown impressive results across a variety of Atari games.

\paragraph{DQN}
For our task, we apply a {\em Deep Q Network\/} (DQN) which uses a deep neural network for approximating the action-value function $Q(s,a;\theta)$, mapping from an input state $s$ and action $a$ to Q values. In particular, two important improvements have been proposed by \cite{mnih2015human} for the learning process, including (1) experience replay, the agent stores observed transitions in a memory buffer for some time, and uniformly samples from the memory to update the network (2) the target network, agent maintains two networks for the loss function --- one for the current estimator of Q function and one for the surrogate of the true Q function. For the current estimator, the parameters are constantly updated. For the surrogate, the parameters are only updated for every certain number of steps from the current estimator network otherwise kept fixed.  

\paragraph{Learning Goal-Parameterized Policies}
To plan with respect to the specific goal, we can parametrize the agent's policy $\pi$ by the goal $g$: 
\begin{equation}
    \pi(s,g,a)
\end{equation}

Since in this work, we applies DQN as value-based method, this corresponds to the update to original Q function with the additional goal information. The new Q-value function is hence defined as:

\begin{equation}
    Q^\pi(s,g,a) = \mathbf{E}[\sum\limits_{i=0}^T \gamma^i r_{t+i+1}|s_t=s,g,a_t=a]
\end{equation}

\begin{figure}
\centering
\includegraphics[width=0.85\linewidth]{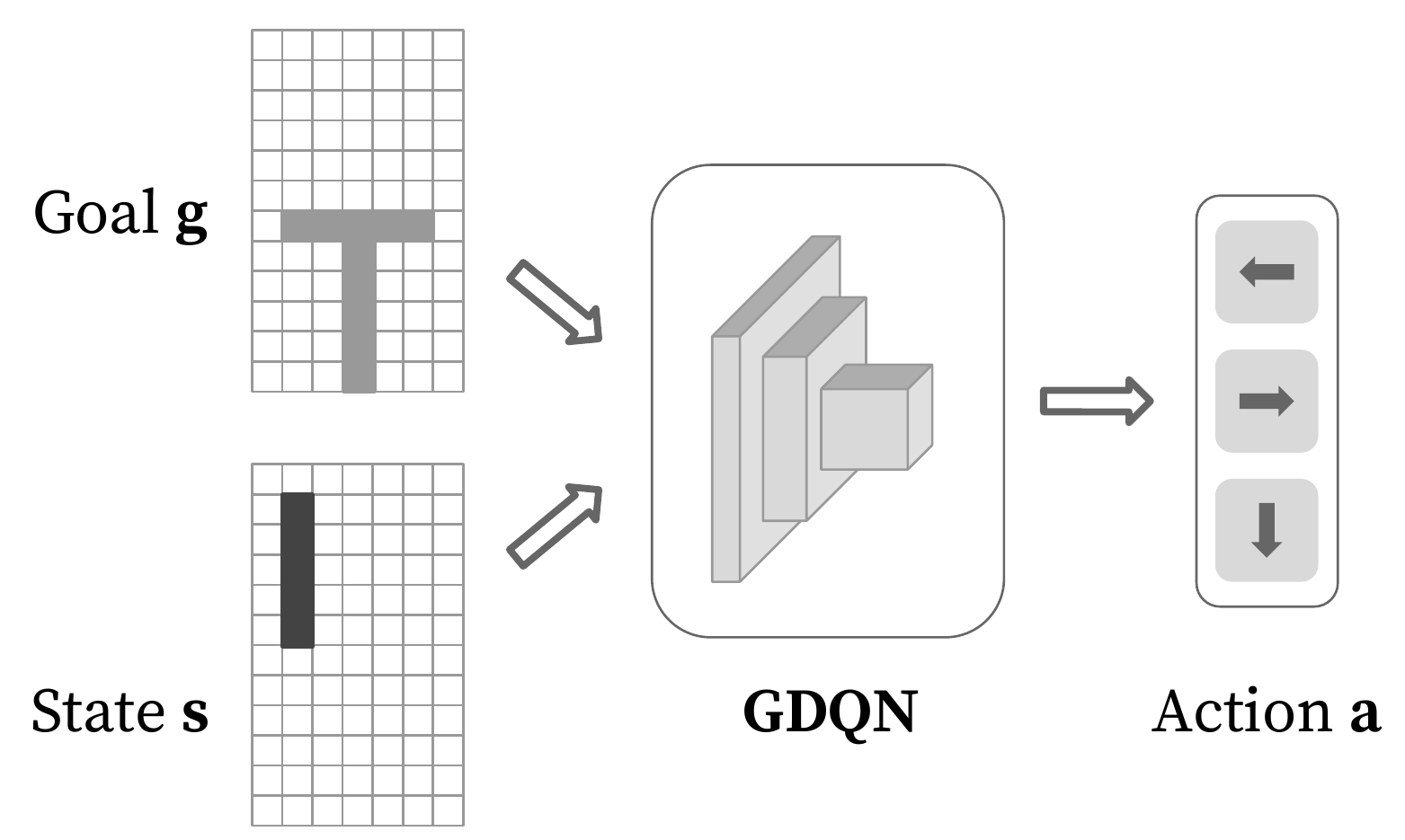}
\caption{Our proposed model GDQN which extends the $Q$-function approximator to integrate goal information.}
\label{fig:goal_net}
\end{figure}

As shown in Figure~\ref{fig:goal_net}, in contrast to the original DQN model, where state and action are used to estimate Q-value,  the new model further include the current goal into the network to produce the estimate. We call this model as Goal-Parametrized Q Network (GDQN). 

The resulted loss function is as:

\begin{equation}
    L_Q = \mathbf{E}[(R + \gamma \text{max}_a' Q^\pi(s',g,a';\theta^-) - Q(s,g,a;\theta))^2 ]
\end{equation}

where $\theta^-$ are the previous parameters and the optimization is with respect to $\theta$. 

\subsubsection{Implementation Details}
The DQN agent is implemented in Theano and Keras to adapt to the settings in our experiment, while we use a $2$ hidden layer (each with $64$ hidden units and rectified linear activation) multilayer perceptron (MLP) for most cases, we additionally swap the MLP with the CNN and follow the reported parameter settings as in the original paper \citep{mnih2015human} to ensure our implementation can reach similar performance.

Note we don't apply the frame-skipping technique \citep{bellemare2012investigating} used for Atari games \citep{mnih2015human} allowing the agent sees and selects actions on every $k$th frame where its last action is repeated on skipped frames. It does not suit our task, in particular when the moving block is getting close to the existing structure, simply repeating action decided from previous frame can cause unintended collision or collapse. 

\paragraph{Reward}
In the target stacking task, the agent gets reward $+1$ when the episode ends with complete reproduction of the target structure, otherwise $0$ reward.

Further, we explore reward shaping \citep{ng1999policy} in the task providing more prompt intermediate reward. Two types of reward shaping are included: overlap ratio and distance transform. 

\begin{figure}
\centering
\begin{subfigure}{0.47\linewidth}
\includegraphics[width=\textwidth]{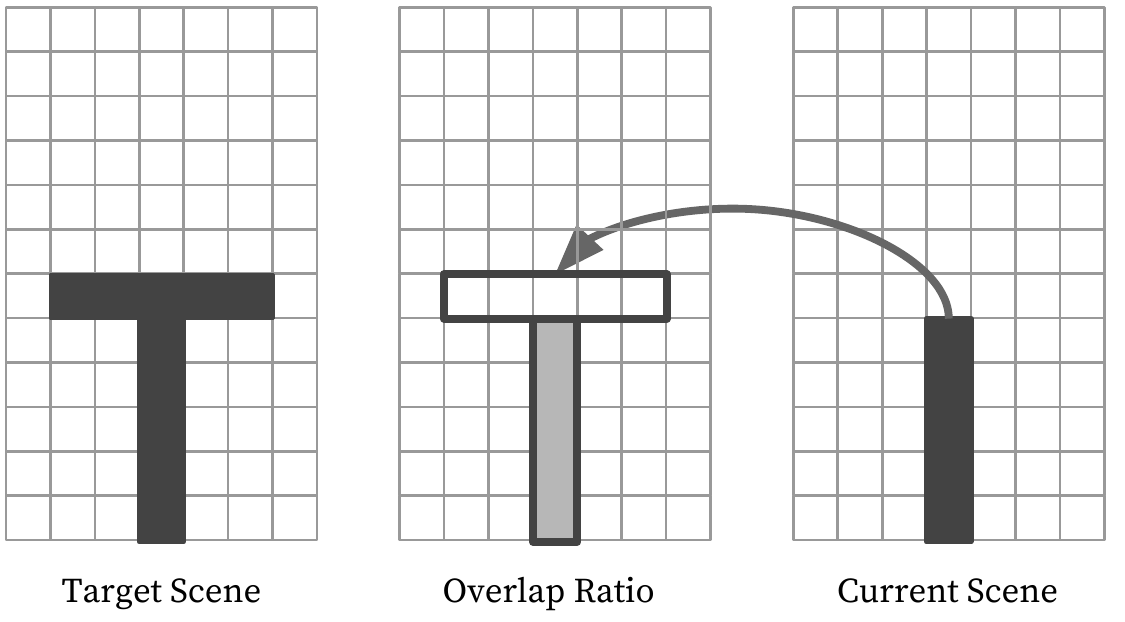}
\caption{}
\label{fig:overlap_ratio}
\end{subfigure}
~
\begin{subfigure}{0.47\linewidth}
\includegraphics[width=\textwidth]{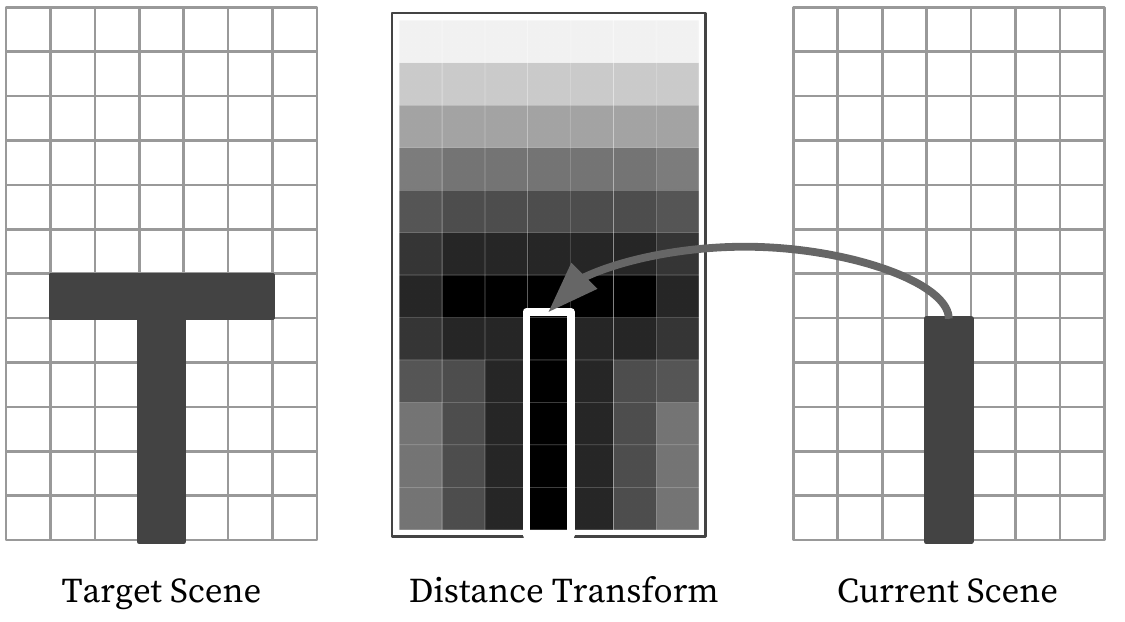}
\caption{}
\label{fig:dist_transform}
\end{subfigure}
\caption{Reward shaping used in target stacking. (\protect\subref{fig:overlap_ratio}): overlap ratio to the target. The gray area in the middle figure denotes the intersected foreground region between current and target scene, and the overlap ratio is the ratio between the areas of the two. (\protect\subref{fig:dist_transform}): distance under the distance transform of the target. The middle figure denotes the distance transform under the target shown in the left. The distance from current scene to the target is the sum of distances masked by the current scene in the distance transform.}
\label{fig:reward_shape}
\end{figure}

For the overlap ratio, for each state $s_t$ under the same target $g_i$, an overlap ratio is counted as the ratio between the intersected foreground region (of the current state and the target state) and the target foreground region (shown in Figure~\ref{fig:overlap_ratio}):

\begin{equation}\label{eq:ovl}
o(s_t,g_i) = \frac{s_t \cap g_i}{g_i}
\end{equation}

For each transition $(s_t, a_t, s_{t+1})$, the reward is defined by the change of overlap ratio before and after the action: 
\begin{equation}
    r_t =
    \begin{cases}
      1, & \text{if}\ \Delta o_{t\rightarrow t+1} = o(s_{t+1}) - o(s_t) >  0 \\
      -1, & \text{if}\ \Delta o_{t\rightarrow t+1} = o(s_{t+1}) - o(s_t) <  0\\
      0, & \text{otherwise}
    \end{cases}
\end{equation}
The intuition is that actions increasing the current state to become more overlapped with the target scene should be encouraged.

For the distance transform \citep{fabbri20082d}, it generates a map $D$ whose value in each pixel $p$ is the smallest distance from it to a target object $O$:

\begin{equation}
D(p) = \text{min}\{\text{dist}(p,q)|q\in O\}
\end{equation}

where $\text{dist}$ can be any valid distance metric, like Euclidean or Manhattan distance. 

For each state $s_t$ under the same target $g_i$, a distance to the goal is the sum of all the element-wise distance in $s_t$ to $g_i$ under $D_{g_i}$ (shown in Figure~\ref{fig:dist_transform}) as: 

\begin{equation}
d(s_t,g_i) = \sum_j D_{g_i}(s_t^j), s_t^j \in s_t
\end{equation}

For each transition $(s_t, a_t, s_{t+1})$, the reward is defined as: 
\begin{equation}
    r_t =
    \begin{cases}
      1, & \text{if}\ \Delta d_{t\rightarrow t+1} = d(s_{t+1}) - d(s_t) <  0 \\
      -1, & \text{if}\ \Delta d_{t\rightarrow t+1} = d(s_{t+1}) - d(s_t) >  0\\
      0, & \text{otherwise}
    \end{cases}
\end{equation}

The intuition behind this is that action decreasing the distance between the current state and the target scene should be encouraged.

\subsubsection{Toy Example with Goal Integration}

\begin{figure}
    \centering
    \begin{subfigure}[b]{0.25\linewidth}
    	\centering
		\includegraphics[height=2.6cm]{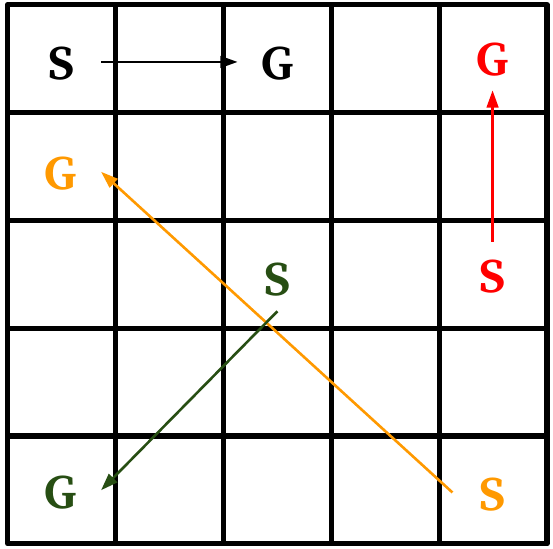}
      	\caption{}
      	\label{fig:nav_sample}
    \end{subfigure}%    
    \begin{subfigure}[b]{0.75\linewidth}
    	\centering
        \begin{tabular}{c|cc}
        \toprule
        Grid Size & DQN  & GDQN \\ 
        \midrule
        $5\times 5$       & 0.67 & 0.97 \\
        $7\times 7$       & 0.67 & 0.95 \\
        \bottomrule
        \end{tabular}
        \caption{}
        \label{tab:gridworld_result}
    \end{subfigure}    
    \caption{\protect\subref{fig:nav_sample}: Navigation task, each color denotes a different episode, per episode, a random pair of starting and goal location are generated, the agent needs to reach the goal. \protect\subref{tab:gridworld_result}: Results from navigation task. }
    \label{fig:nav}
\end{figure}

At first, we introduce a type of navigation task in the classic gridworld environment. 
As shown in Figure~\ref{fig:nav_sample}, for each episode, we generate a random start and goal point in the grid world whereas the start and the goal have to be different from each other, the agent has to find the way to this randomly generated goal. The agent is rewarded only when it reaches it. The agent needs to reach the goal with four possible actions $\{\text{left},\text{right},\text{up},\text{down}\}$. Action that will make the agent go off the grid will leave it stay in the same location. The episode only terminates once the agent reaches the goal. The agent only receive reward $+1$ when reaching the current goal. Two different sizes of gridworld are tested at $5\times5$ and $7\times7$. We learn a single model for each size of the mazes.

The training epoch size is $1000$ in steps for the smaller gridworld and $3000$ for the larger one, the test sizes are the same for both at $100$. All the agents run for $100$ epochs and the $\epsilon$ for $\epsilon$-greedy anneals linearly from $1.0$ to $0.1$ over the first $20$ epochs, and fixed at $0.1$ thereafter. The memory buffer size is set the same to the annealing length, i.e. for the smaller gridworld, the buffer size equals to the length $20$ epochs in training with $20000$ steps whereas for the larger one, the buffer size is $30000$ steps.
We measure the proportion of episodes in the test epoch that reaches the goal in shortest distance as the success ratio. The results are shown in Table~\ref{tab:gridworld_result} for the best agents throughout the training process.

As in this simple task with relative small state space, DQN gets some performance due to running an average policy across all the the goals, but this is not addressing the task we set out to do. In contrast, GDQN parametrized specifically to include goal information achieves significant better results on both sizes of the environment.

\subsubsection{Target Stacking}

\begin{table*}
\centering
\begin{tabular}{@{}clccccccccccc@{}}
\toprule
\multicolumn{2}{c}{\multirow{2}{*}{Num. of Blks.}} & \multicolumn{2}{c}{DQN} &  & \multicolumn{2}{c}{GDQN} &  & \multicolumn{2}{c}{GDQN + OR} &  & \multicolumn{2}{c}{GDQN + DT} \\ \cmidrule(lr){3-4} \cmidrule(lr){6-7} \cmidrule(lr){9-10} \cmidrule(l){12-13} 
\multicolumn{2}{c}{}                               & OR         & SR         &  & OR          & SR         &  & OR            & SR            &  & OR            & SR            \\ \midrule
\multicolumn{2}{c}{2}                              & 0.70          & 0.70          &  & 0.82           & 0.72          &  & 0.84             & 0.77             &  & {\bf 0.88}             & {\bf 0.78}             \\
\multicolumn{2}{c}{3}                              & 0.43          & 0.43          &  & 0.76           & {\bf 0.67}          &  & {\bf 0.86}             & 0.63             &  & 0.83             & 0.65             \\
\multicolumn{2}{c}{4}                              & 0.03          & 0.0          &  & 0.41           & 0.17          &  & 0.73             & 0.55             &  & {\bf 0.79}             & {\bf 0.56}             \\ \bottomrule
\end{tabular}
\caption{Results for target stacking. For ``GDQN + X'', X denotes different ways for reward shaping as described in previous section, OR for overlap ratio, DT for distance transform. For metrics, OR stands for average overlap ratio, SR for average success rate.}
\label{tab:tgt_stack}
\end{table*}

\begin{figure}
    \centering
    \begin{subfigure}[b]{0.42\linewidth}
        \includegraphics[width=\textwidth]{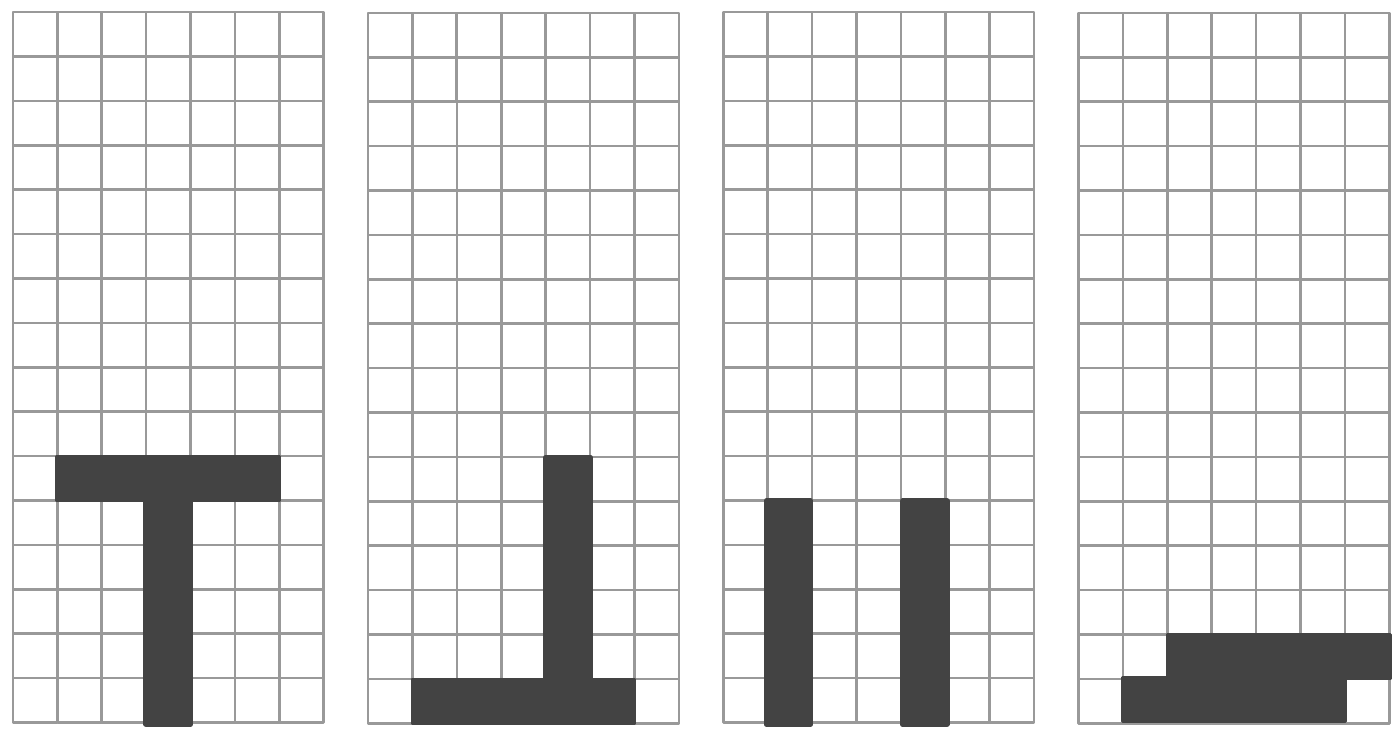}
        \caption{}
        \label{fig:2blks_tgt_stack}
    \end{subfigure}
    ~ 
    \begin{subfigure}[b]{0.42\linewidth}
        \includegraphics[width=\textwidth]{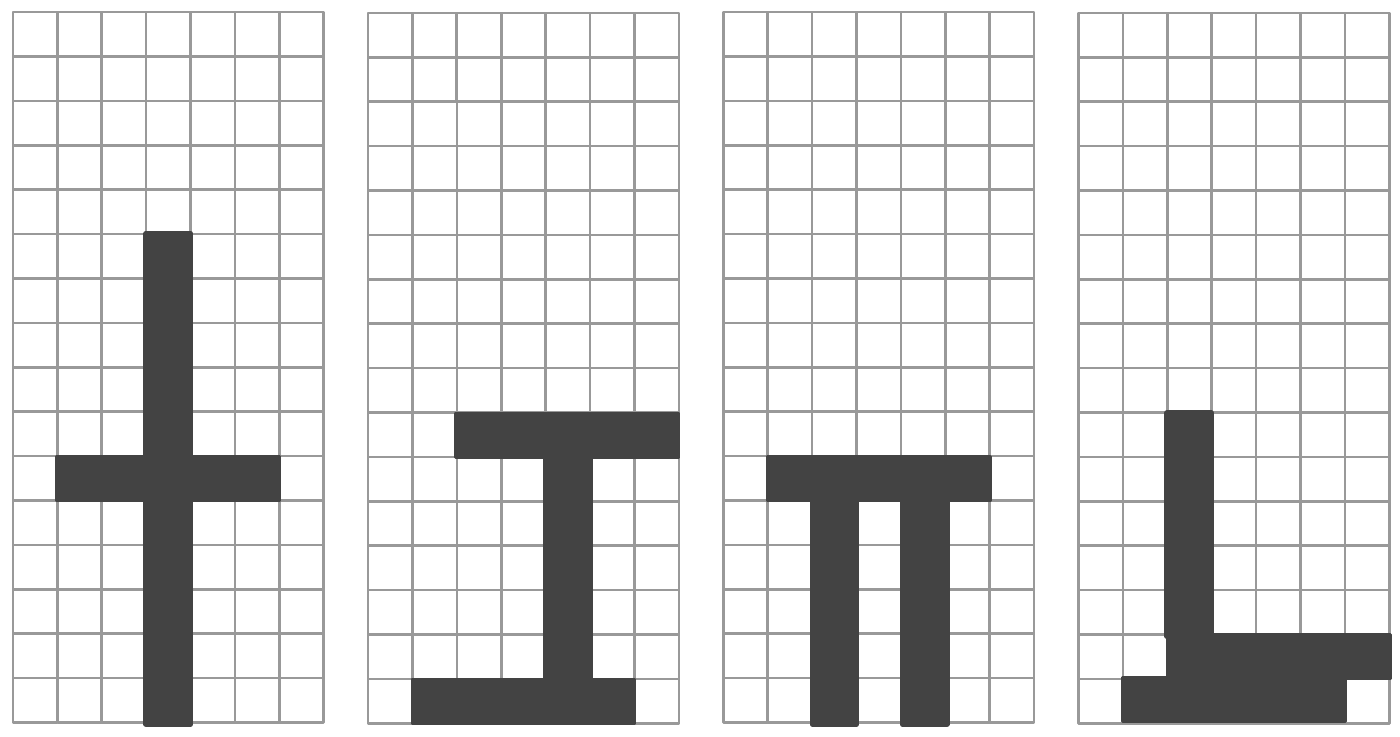}
        \caption{}
        \label{fig:3blks_tgt_stack}
    \end{subfigure}
    ~ 
    \begin{subfigure}[b]{0.42\linewidth}
        \includegraphics[width=\textwidth]{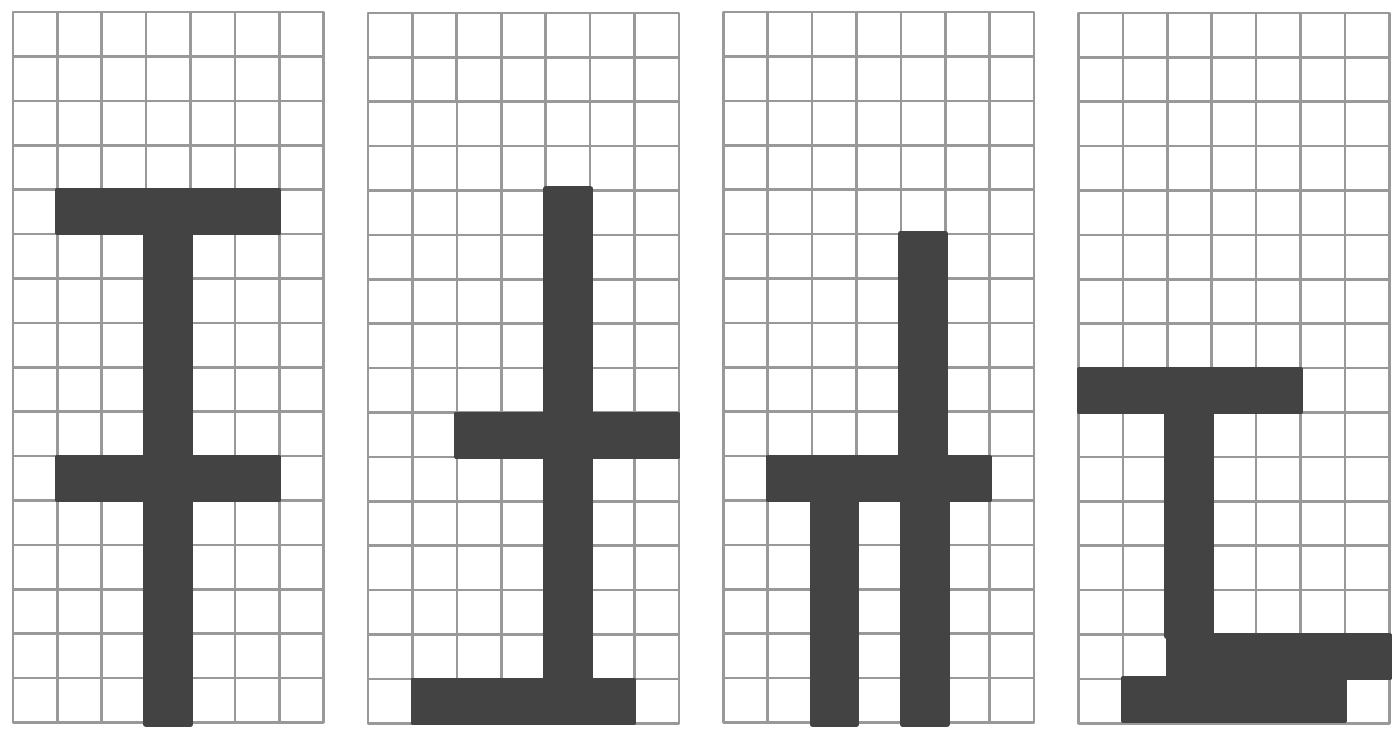}
        \caption{}
        \label{fig:4blks_tgt_stack}
    \end{subfigure}
    \caption{\protect\subref{fig:2blks_tgt_stack}: Targets for $2$ blocks.\protect\subref{fig:3blks_tgt_stack}: Targets for $3$ blocks. \protect\subref{fig:4blks_tgt_stack}: Targets for $4$ blocks.}
    \label{fig:tgt_stack}
\end{figure}

We set up $3$ groups of target structures consisted of different number of blocks $\{2,3,4\}$ in the scene as shown in Figure~\ref{fig:tgt_stack}. Within each group of target shapes, a random target (with the accompanied orientation order) is first picked at the very beginning for individual episode and then each block is spawned at randomized location as described earlier in the target stacking environment. Each training epoch consists of $10000$ steps and each test epoch with $1000$ steps. Similar to the setting in the toy example, all the agents run for $100$ epochs and the $\epsilon$ anneals for the first $20$ epochs, and the memory buffer size is set as long as the annealing steps at $200K$ steps. 

We computed both average overlap ratio (OR) and success rate (SR) for the finished stacking episodes in each test epoch. Here overlap ratio is the same as defined in the reward shaping in Equation~\ref{eq:ovl}, but simply measures the end scene over the assigned target scene. This tells the relative completion of the stacked structure in comparison to the assigned target structure, the higher the value is, the better completion it is to the target. At the maximum of $1$, it suggests completely reproduction of the target. The success rate counts the ratio how many episodes complete the exact same shape as assigned over the total number of episodes finished in the test epoch. This is the absolute metric counting overall successful stacking. The results are shown in Table~\ref{tab:tgt_stack} for the best agents throughout the training process.

Over all groups on both metrics, we observe GDQN outperforms DQN, showing the importance of integrating goal information. In general, the more blocks in the task, the more difficult it becomes. When there are only small number of blocks ($2$ blocks and $3$ blocks) in the scene, the single policy learned by DQN averages over the few target shapes can still work to some extent. However when introducing more blocks into the scene, it becomes more and more difficult for this averaged model to handle. As we can see from the result, there is already a significant decrease of performance (success rate drops from $0.70$ to $0.43$) when increasing the blocks number from $2$ to $3$, whereas GDQN's performance only decreases slightly from $0.72$ to $0.67$. In $4$ blocks scene, the DQN can no longer reproduce any single target ($0.0$ for success rate, $0.03$ for overlap ratio) while GDQN parametrized specifically to include goal information can still do. Though the success rate (absolute completion to the target) for the basic GDQN is relatively low at $0.17$ but the average overlap ratio (relative completion to the target) still holds up pretty well at $0.41$. Also we see reward shaping can further improves GDQN model, in particular distance transform can boost the performance more than overlap ratio.

\begin{figure}
\addtolength{\tabcolsep}{-3.5pt}
\begin{center}$
\begin{tabular}{cccc}
\label{fig:hstack1}\includegraphics[width=0.17\linewidth]{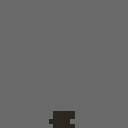}&
\label{fig:hstack2}\includegraphics[width=0.17\linewidth]{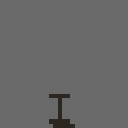}&
\label{fig:hstack3}\includegraphics[width=0.17\linewidth]{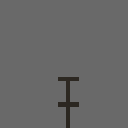}&
\label{fig:hstack4}\includegraphics[width=0.17\linewidth]{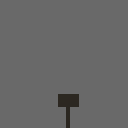}\\
\end{tabular}$
\end{center}
\addtolength{\tabcolsep}{3.5pt}
\vspace{-0.4cm}
\caption{Example scenes constructed by the agent.}
\label{fig:scene_sample}
\end{figure}
\vspace{-0.4cm}
\section{Conclusion}
In this work, we answer the question if and how well we can build up a machinery to learn a manipulation task under physics constraints with visual perception. In particular, We focus on a wood block stacking problem and explored frameworks from two inter-related but different perspectives.

In the first part of the work, we aim to guide the robot to stably place a single block in the scene. To achieve this, we envision a mechanism of explicit physics understanding to predict physical stability directly from visual input, bypassing explicit 3D representations and physical simulation. We initially evaluate our model on a synthetic dataset, covering a range of conditions including variations in number of blocks, size of blocks and 3D structure of the overall tower. The results reflect the challenges of inference with growing complexity of the structure. To further understand the results, we conduct a human subject study on a subset of our synthetic data and show that our model achieves comparable or even better results than humans in the same setting. Moreover, we investigate the discriminative image regions found by the model and spot correlation between such regions and initial collapse area in the structure. Finally, We apply our approach to a block stacking setting and show that our model successfully guide a robot for placements of new blocks by predicting the stability of future states.

In the following part of the work, we explore further on a more challenging target stacking task where the agent stacks blocks to reproduces a tower shown in an image. It presents a distinct type of challenge requiring the agent to reach a given goal state that may be different for every new trial. To this end, we propose an alternative framework where the manipulation is learned end-to-end through pure trial and error, bypassing the explicitly modeling for the corresponding physics knowledge as we did in the previous task. We create a synthetic block stacking environment with physics simulation in which the agent can interactively stack the block. In particular, we propose a goal-parametrized GDQN model to plan with respect to the specific goal, allowing better generalization across different goals. We validate the model on both a navigation task in a classic gridworld environment with different start and goal positions and the block stacking task itself with different target structures.

\bibliographystyle{SageH}
\bibliography{mybib}

\end{document}